\documentclass[journal]{IEEEtran}
\newcommand{\ignore}[1]{}
\usepackage[utf8]{inputenc}
\usepackage[T1]{fontenc}
\usepackage{amsmath,epsfig}
\usepackage{graphicx}
\usepackage{bm}
\usepackage{amssymb}
\usepackage{cite}
\usepackage{array}
\usepackage{extarrows}
\usepackage{url}
\usepackage{float}

\usepackage{color}

\usepackage{multirow}
\usepackage{booktabs}
\usepackage{blkarray}
\usepackage{multirow}
\usepackage{array}
\usepackage{color}
\usepackage{pmat}

\begin{document}
%
\title{TransY-Net: Learning Fully Transformer Networks for Change Detection of Remote Sensing Images}
\author{
        Tianyu Yan,
        Zifu~Wan,
        Pingping~Zhang$^{*}$,
        Gong Cheng
        and~Huchuan~Lu
\thanks{

This work was in part sponsored by CAAI-Huawei MindSpore Open Fund under Grant No. CAAIXSJLJJ-2021-067A, and the National Natural Science Foundation of China under Grant No. 62101092.

($^{*}$Corresponding author: Pingping Zhang.)

TY. Yan, ZF. Wan, PP. Zhang are with School of Artificial Intelligence, Dalian University of Technology, Dalian, 116024, China. (Email: 2981431354@mail.dlut.edu.cn;2537998622@qq.com;zhpp@dlut.edu.cn)

G. Cheng is with School of Automation, Northwestern Polytechnical University, Xi'an, 710072, China. (Email: gcheng@nwpu.edu.cn)

HC. Lu is with School of Information and Communication Engineering, Dalian University of Technology, Dalian, 116024, China.  (Email: lhchuan@dlut.edu.cn)
}
}
\maketitle
\markboth{IEEE Transactions on Geoscience and Remote Sensing}{}
\begin{abstract}
In the remote sensing field, Change Detection (CD) aims to identify and localize the changed regions from dual-phase images over the same places.
Recently, it has achieved great progress with the advances of deep learning.
However, current methods generally deliver incomplete CD regions and irregular CD boundaries due to the limited representation ability of the extracted visual features.
To relieve these issues, in this work we propose a novel Transformer-based learning framework named TransY-Net for remote sensing image CD, which improves the feature extraction from a global view and combines multi-level visual features in a pyramid manner.
More specifically, the proposed framework first utilizes the advantages of Transformers in long-range dependency modeling.
It can help to learn more discriminative global-level features and obtain complete CD regions.
Then, we introduce a novel pyramid structure to aggregate multi-level visual features from Transformers for feature enhancement.
The pyramid structure grafted with a Progressive Attention Module (PAM) can improve the feature representation ability with additional inter-dependencies through spatial and channel attentions.
Finally, to better train the whole framework, we utilize the deeply-supervised learning with multiple boundary-aware loss functions.
Extensive experiments demonstrate that our proposed method achieves a new state-of-the-art performance on four optical and two SAR image CD benchmarks.
The source code is released at https://github.com/Drchip61/TransYNet.
\end{abstract}
\IEEEpeerreviewmaketitle
\section{Introduction}
\IEEEPARstart{C}hange Detection (CD) plays an important role in the field of remote sensing.
It aims to detect the key change regions in dual-phase remote sensing images captured at different times but over the same scene area.
In fact, remote sensing image CD has been used in many real-world applications, such as land-use planning, urban expansion management, geological disaster monitoring, ecological environment protection.
However, due to the change regions can be any shapes in complex scenarios, there are still many challenges for high-accuracy CD.
In addition, remote sensing image CD by handcrafted methods is time-consuming and labor-intensive, thus there is a great need for fully-automatic and highly-efficient CD.

In recent years, deep learning has been widely used in remote sensing image processing due to its powerful feature representation capabilities, and has shown great potential in CD.
With Convolutional Neural Networks (CNN)~\cite{he2016deep} and Transformers~\cite{dosovitskiy2020image,liu2021swin}, many CD methods extract more discriminative features and have demonstrated good performances.
However, previous CD methods still have the following shortcomings: 1) With the resolution improvement of remote sensing images, rich semantic information contained in high-resolution images is not fully utilized. As a result, current CD methods are unable to distinguish pseudo changes such as shadow, vegetation and sunshine in sensitive areas. 2) Boundary information in complex remote sensing images is often missing. In previous methods, the extracted changed areas often have regional holes and their boundaries can be very irregular, resulting in a poor visual effect~\cite{liu2020building}. 3) The temporal information contained in dual-phase remote sensing images is not fully utilized, which is also one of the reasons for the low performance of current CD methods.

To tackle above issues, in this work we propose a novel Transformer-based learning framework named TransY-Net for remote sensing image CD, which improves the feature extraction from a global view and combines multi-level visual features in a pyramid manner.
More specifically, the proposed framework has a Y-shape structure whose input is a dual-phase remote sensing image pair.
We first utilize the advantages of Transformers in long-range dependency modeling to learn more discriminative global-level features.
Then, to highlight the change regions, the summation features and difference features are generated by directly comparing the temporal features of dual-phase remote sensing images.
Thus, one can obtain complete CD regions.
To improve the boundary perception ability, we further introduce a pyramid structure to aggregate multi-level visual features from Transformers.
The pyramid structure grafted with a Progressive Attention Module (PAM) can improve the feature representation ability with additional inter-dependencies through spatial and channel attentions.
Finally, to better train the framework, we utilize the deeply-supervised learning with multiple boundary-aware loss functions.
Extensive experiments show that our method achieves a new state-of-the-art performance on four optical and two SAR image CD benchmarks.

The main contributions are summarized as follows:
\begin{itemize}
\item We propose a Transformer-based learning framework (\emph{i.e.}, TransY-Net) for remote sensing image CD, which can improve the feature extraction from a global view and combine multi-level visual features in a pyramid manner.
\item We propose a novel pyramid structure grafted with a Progressive Attention Module (PAM) to further improve the feature representation ability with additional inter-dependencies through spatial and channel attentions.
\item We introduce the deeply-supervised learning with multiple boundary-aware loss functions, to address the irregular boundary problem in CD.
\item Extensive experiments on four optical and two SAR image CD benchmarks show that our framework attains better performances than most state-of-the-art methods.
\end{itemize}

We note that this work is an extension of its previous conference version~\cite{yan2022fully} with some key improvements as follows:
1) We propose a more powerful PAM with joint spatial and channel attentions.
2) We enhance the multi-level visual features with multi-scale pooling.
3) We provide more discussions with other Transformer-based methods.
4) We add more experimental results to verify the effectiveness of the proposed framework and modules.
\section{Related Work}
\subsection{Change Detection of Remote Sensing Images}
Technically, the task of change detection takes dual-phase remote sensing images as inputs, and predicts the change regions of the same places.
Before deep learning, direct classification-based methods witness the great progress in CD.
For example, Change Vector Analysis (CVA)~\cite{xiaolu2011change} is powerful in extracting pixel-level features and is widely utilized in CD.
With the rapid improvement in image resolution, more details of objects have been recorded in remote sensing images.
Therefore, many object-aware methods are proposed to improve the CD performance.
For example, Tang \emph{et al.}~\cite{tang2011object} propose an object-oriented CD method based on the Kolmogorov--Smirnov test.
Li \emph{et al.}~\cite{li2016change} propose the object-oriented CVA to reduce the number of pseudo detection pixels.
With multiple classifiers and multi-scale uncertainty analysis, Tan \emph{et al.}~\cite{tan2019object} build an object-based approach for complex scene CD.
Although above methods can generate CD maps from dual-phase remote sensing images, they generally deliver incomplete CD regions and irregular CD boundaries due to the limited representation ability of the extracted visual features.

With the advances of deep learning, many works improve the CD performance by extracting more discriminative features.
For example, Zhang~\emph{et al.}~\cite{zhang2016feature} utilize a Deep Belief Network (DBN) to extract deep features and represent the change regions by patch differences.
Saha~\emph{et al.}~\cite{saha2019unsupervised} combine a pre-trained deep CNN and traditional CVA to generate certain change regions.
Hou~\emph{et al.}~\cite{hou2017change} take the advantages of deep features and introduce the low rank analysis to improve the CD results.
Peng~\emph{et al.}~\cite{peng2019unsupervised} utilize saliency detection analysis and pre-trained deep networks to achieve unsupervised CD.
Since change regions may appear in any places, Lei~\emph{et al.}~\cite{lei2019multiscale} integrate Stacked Denoising AutoEncoders (SDAE) with multi-scale superpixel segmentation to realize superpixel-based CD.
Similarly, Lv~\emph{et al.}~\cite{lv2018deep} utilize a Stacked Contractive AutoEncoder (SCAE) to extract temporal change features from image superpixels, then adopt a clustering method to produce accurate CD maps.

Meanwhile, some methods formulate the CD task as a binary image segmentation task.
Thus, CD can be finished in a supervised manner.
For example, Alcantarilla~\emph{et al.}~\cite{alcantarilla2018street} first concatenate dual-phase images as one image with six channels.
Then, the six-channel image is fed into a Fully Convolutional Network (FCN) to realize the CD.
Similarly, Peng~\emph{et al.}~\cite{peng2019end} combine bi-temporal remote sensing images as one input for CD.
Daudt~\emph{et al.}~\cite{daudt2018fully} utilize Siamese networks to extract features for each remote sensing image, then predict the CD maps with fused features.
The experimental results prove the efficiency of Siamese networks.
Furthermore, Guo~\emph{et al.}~\cite{guo2018learning} use a fully convolutional Siamese network with a contrastive loss to measure the change regions.
Zhang~\emph{et al.}~\cite{zhang2020deeply} propose a deeply-supervised image fusion network for CD.
There are also some works focused on specific object CD.
For example, Liu~\emph{et al.}~\cite{liu2020building} propose a dual-task constrained deep Siamese convolutional network for building CD.
Jiang~\emph{et al.}~\cite{jiang2020pga} propose a pyramid feature-based attention-guided Siamese network for building CD.
Lei~\emph{et al.}~\cite{lei2020hierarchical} propose a hierarchical paired channel fusion network for street scene CD.
The aforementioned methods have shown great success in feature learning for CD.
However, these methods have limited global representation capabilities and usually focus on local regions of changed objects.
We find that Transformers have strong characteristics in extracting global features.
Thus, different from previous works, we take the advantages of Transformers, and propose a new learning framework for more discriminative feature representations.
\begin{figure*}
\centering
\resizebox{0.9\textwidth}{!}
{
\begin{tabular}{@{}c@{}c@{}}
\includegraphics[width=1\linewidth,height=0.4\linewidth]{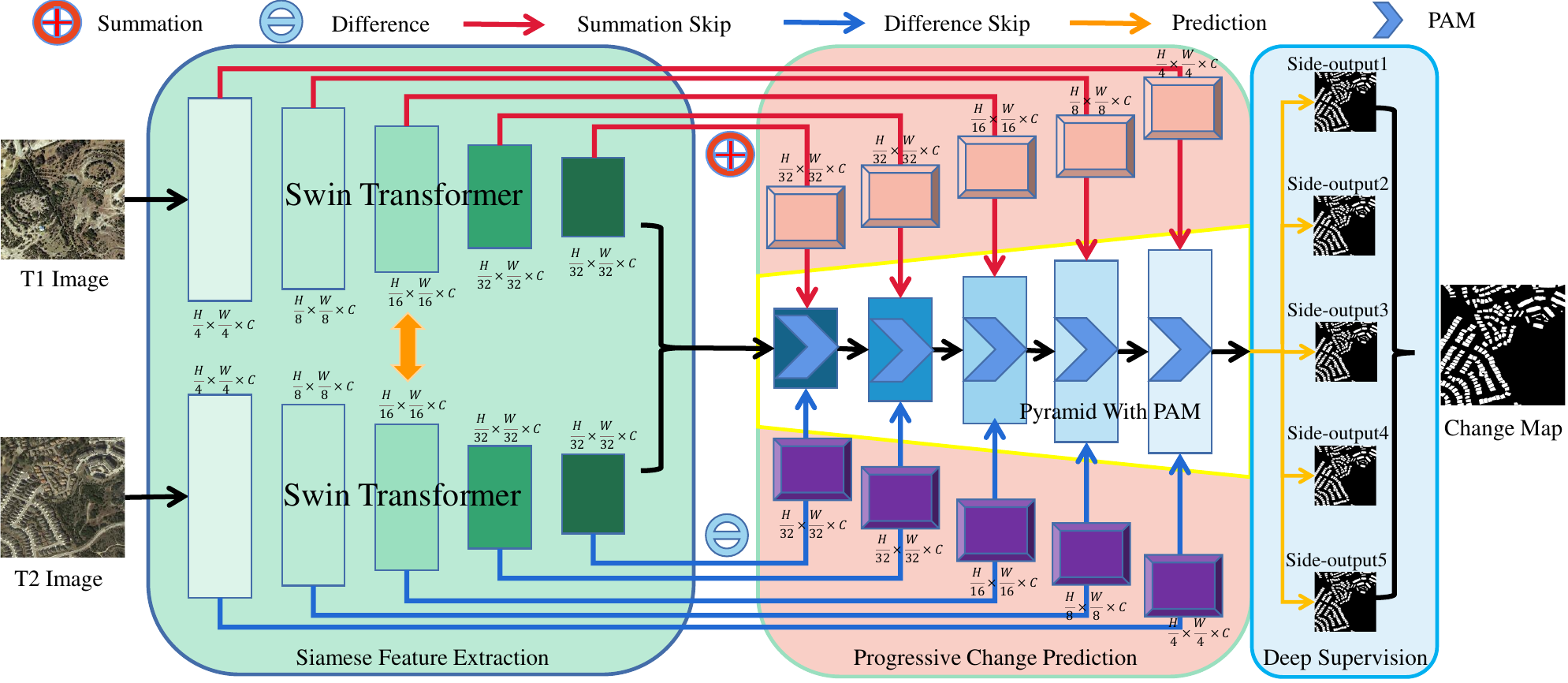} \\
\end{tabular}
}
\caption{The overall structure of our proposed framework (TransY-Net).
Firstly, a typical Siamese Feature Extraction (SFE) network with shared Swin Transformers are utilized to extract multi-level feature maps from dual-phase remote sensing images.
Then, the Deep Feature Enhancement (DFE) is introduced to highlight the change regions with summation features and difference features.
Afterwards, the Progressive Change Prediction (PCP) is adopted to encode and integrate multi-level features progressively for the final change map prediction.
To improve the representation ability, a pyramid structure with a Progressive Attention Module (PAM) is utilized with additional interdependencies through spatial and channel attentions.
Finally, the Deep Supervison (DS) is utilized with multiple boundary-aware loss functions to train the whole framework.
}
\label{fig:Framework}
\vspace{-4mm}
\end{figure*}
\subsection{Vision Transformers for Change Detection}
Transformers~\cite{vaswani2017attention} are firstly proposed for time series tasks, such as natural language processing, speech generation.
Recently, they have been applied to many computer vision tasks, such as image classification~\cite{dosovitskiy2020image,liu2021swin}, person re-identification~\cite{zhang2021hat,liu2021video} and so on.
Inspired by the extreme effectiveness, Zhang~\emph{et al.}~\cite{zhang2022swinsunet} deploy a Swin Transformer structure~\cite{liu2021swin} with a U-Net~\cite{ronneberger2015u} for remote sensing image CD.
Zheng~\emph{et al.}~\cite{zheng2022changemask} design a deep Multi-task Encoder-Transformer-Decoder (METD) architecture for semantic CD.
Wang~\emph{et al.}~\cite{wang2021transcd} incorporate a Siamese Vision Transformer (SViT) into a feature difference framework for CD.
To take the advantages of both Transformers and CNNs, Wang~\emph{et al.}~\cite{wang2022network} propose to combine a Transformer and a CNN for remote sensing image CD.
Li~\emph{et al.}~\cite{li2022transunetcd} propose an encoding-decoding hybrid framework for CD, which has the advantages of both Transformers and U-Net.
Bandara~\emph{et al.}~\cite{bandara2022transformer} unify hierarchically structured Transformer encoders with Multi-Layer Perception (MLP) decoders in a
Siamese network to efficiently render multi-scale long-range details for accurate CD.
Chen~\emph{et al.}~\cite{chen2021remote} propose a Bitemporal Image Transformer (BIT) to efficiently and effectively model contexts within the spatial-temporal domain for CD.
Ke~\emph{et al.}~\cite{ke2022hybrid} propose a hybrid Transformer with token aggregation for remote sensing image CD.
Song~\emph{et al.}~\cite{song2022mstdsnet} combine the multi-scale Swin Transformer and a deeply-supervised network for CD.
All these methods have shown that Transformers can model the inter-patch relations for strong feature representations.
However, these CD methods do not take the full abilities of Transformers in multi-level feature learning.
Different from existing Transformer-based CD methods, our proposed approach handles incomplete CD regions and irregular CD boundaries.
Besides, we utilize a Siamese structure to process dual-phase remote sensing images, and introduce a pyramid structure to aggregate multi-level features from Transformers for feature enhancement.
\subsection{Feature Pyramid Methods in Remote Sensing}
Multi-scale features play an important role in remote sensing image processing, including change detection.
As a typical multi-scale feature fusion method, Feature Pyramid Network (FPN)~\cite{lin2017feature} is first proposed for object detection from natural images, and it can aggregate multi-scale features in a coarse-to-fine manner.
Recently, it also shows great successes in remote sensing tasks.
For example, Yang \emph{et al.}~\cite{yang2018automatic} propose a multi-scale rotation dense FPN for automatic ship detection.
Li \emph{et al.}~\cite{li2020radet} refine the FPN and multi-layer attention network for arbitrary-oriented object detection of remote sensing images.
Shamsolmoali \emph{et al.}~\cite{shamsolmoali2021multipatch} utilize a multi-patch FPN for weakly supervised object detection in optical remote sensing images.
Wang \emph{et al.}~\cite{wang2021enhanced} enhance the FPN with deep semantic embedding for remote sensing scene classification.
Gao \emph{et al.}~\cite{gao2018end} combine multiple FPNs for end-to-end road extraction.
Zhang \emph{et al.}~\cite{zhang2021laplacian} introduce a Laplacian FPN for small object detection.
Zhang \emph{et al.}~\cite{chen2021water} combine a FPN and pixel pair matching for water-body segmentation.
We note that our work is indeed inspired by the classical FPN.
However, our work introduces advanced Transformers into the FPN, which can capture more long-range contextual information.
Besides, our pyramid structure can improve the feature representation ability with additional inter-dependencies through spatial and channel attentions.
They are different from the classical FPN.
\section{Proposed Approach}
As shown in Fig.~\ref{fig:Framework}, our proposed framework (TransY-Net) has a Y-shape structure, and includes four key components, \emph{i.e.}, Siamese Feature Extraction (SFE), Deep Feature Enhancement (DFE), Progressive Change Prediction (PCP) and Deep Supervision (DS).
By taking dual-phase images as inputs, SFE first extracts multi-level visual features through two weight-shared Swin Transformers.
Then, DFE utilizes the multi-level visual features to generate summation features and difference features, which highlight the change regions with temporal information.
Afterwards, by integrating all above features, PCP introduces a pyramid structure grafted with a Progressive Attention Module (PAM) for the final CD prediction.
Finally, to train our proposed framework, DS is introduced to achieve the deeply-supervised learning with multiple boundary-aware loss functions for each feature level.
We will elaborate on these key modules in the following subsections.
\subsection{Siamese Feature Extraction}
\begin{figure}
\centering
\resizebox{0.34\textwidth}{!}
{
\begin{tabular}{@{}c@{}c@{}}
\includegraphics[width=1\linewidth,height=1\linewidth]{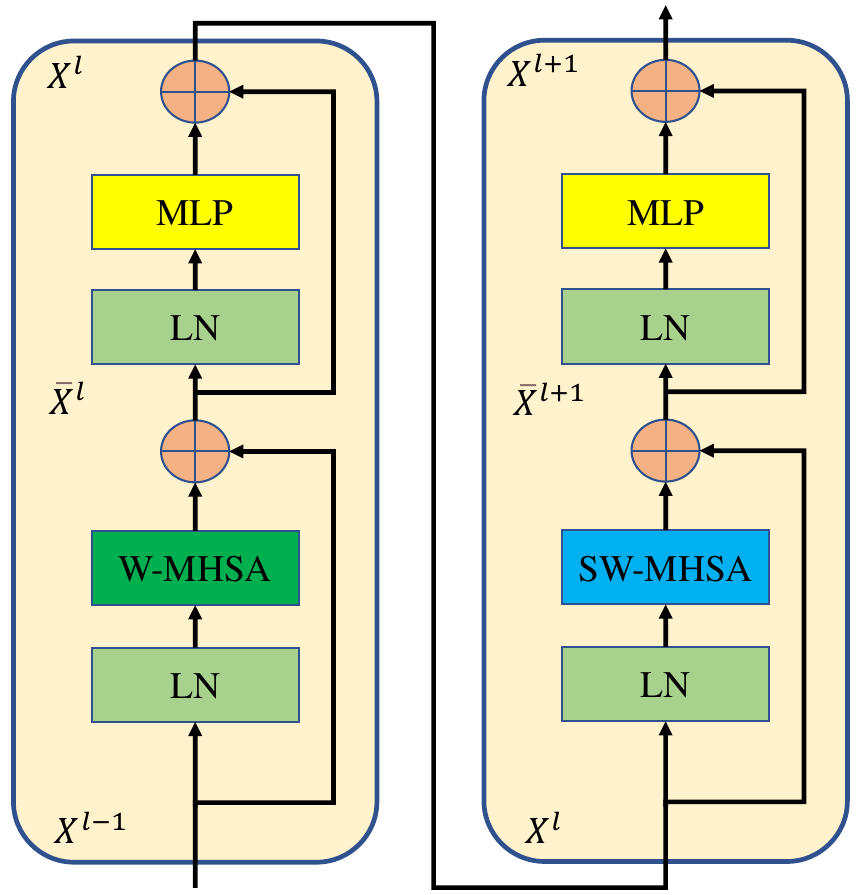} \\
\end{tabular}
}
\caption{The basic structure of the used Swin Transformer block.}
\label{fig:SwinT}
\end{figure}

Following previous works, we introduce a Siamese structure to extract multi-level features from the dual-phase remote sensing images.
More specifically, the Siamese structure contains two encoder branches, which share learnable weights and are used for the multi-level feature extraction of remote sensing images at temporal phase 1 (T1) and temporal phase 2 (T2), respectively.
As shown in the left part of Fig.~\ref{fig:Framework}, we take the Swin Transformer~\cite{liu2021swin} as the basic backbone of the Siamese structure, which involves five stages in total.
Different from other typical Transformers~\cite{vaswani2017attention,dosovitskiy2020image}, the Swin Transformer replaces the standard Multi-Head Self-Attention (MHSA) with Window-based Multi-Head
Self-Attention (W-MHSA) and Shifted Window-based Multi-Head
Self-Attention (SW-MHSA), to reduce the computational complexity of the global self-attention.
To improve the representation ability, the Swin Transformer also introduces the Multi-Layer Perception (MLP), LayerNorm (LN) layers and residual connections.
Fig.~\ref{fig:SwinT} shows the basic structure of the Swin Transformer block used in this work.
Technically, the calculation formulas of all the procedures are given as follows:
\begin{equation}\label{1}
\bar{\textbf{X}}^l = \text{W-MHSA}(\text{LN}(\textbf{X}^{l-1}))+\textbf{X}^{l-1},
\end{equation}
\begin{equation}\label{2}
\textbf{X}^l = \text{MLP}(\text{LN}(\bar{\textbf{X}}^{l}))+\bar{\textbf{X}}^{l},
\end{equation}
\begin{equation}\label{3}
\bar{\textbf{X}}^{l+1} = \text{SW-MHSA}(\text{LN}(\textbf{X}^{l}))+\textbf{X}^{l},
\end{equation}
\begin{equation}\label{4}
\textbf{X}^{l+1} = \text{MLP}(\text{LN}(\bar{\textbf{X}}^{l+1}))+\bar{\textbf{X}}^{l+1},
\end{equation}
where $\bar{\textbf{X}}$ is the output with the W-MHSA or SW-MHSA module and $\textbf{X}$ is the output with the MLP module.
At each stage of the original Swin Transformers, the feature resolution is halved, while the channel dimension is doubled.
More specifically, the feature resolution is reduced from $(H/4)\times(W/4)$ to $(H/32)\times(W/32)$, and the channel dimension is increased from $C$ to $8C$.
In order to take advantages of global-level information, we introduce an additional Swin Transformer block to enlarge the receptive field of the feature maps.
Besides, to reduce the computation, we uniformly change the channel dimension to $C$, and generate encoded features $[\textbf{E}^{1}_{T1}, \textbf{E}^{2}_{T1},...,\textbf{E}^{5}_{T1}]$ and $[\textbf{E}^{1}_{T2}, \textbf{E}^{2}_{T2},...,\textbf{E}^{5}_{T2}]$ for the T1 and T2 images, respectively.
Based on the weight-shared Swin Transformers, the multi-level features can be extracted.
In general, features in the high-level capture global semantic information, while features in the low-level retain local detail information.
Both of them help the detection of change regions.
\subsection{Deep Feature Enhancement}
In complex scenarios, there are many visual challenges for remote sensing image CD.
Thus, only depending on the above features is not enough.
To highlight the change regions, we propose to enhance the multi-level visual features with feature summation and difference, as shown in the top part and bottom part of Fig.~\ref{fig:Framework}.
Here, we note that the difference operation is a typical method for highlighting the changed regions. While the summation operation is also a useful method for feature fusion. When using the summation of two-stream features, the common information is enhanced. It is very useful for the change detection as verified in~\cite{zhang2023escnet}.
More specifically, we first perform a point-wise feature summation and difference, then introduce a contrast feature associated to
each local feature.
The enhanced features can be represented as:
\begin{equation}\label{5}
\bar{\textbf{E}}^{k}_S = \text{ReLU}(\text{BN}(\text{Conv}(\textbf{E}^{k}_{T1}+\textbf{E}^{k}_{T2}))),
\end{equation}
\begin{equation}\label{6}
\bar{\textbf{E}}^{k,m}_{SC} = \bar{\textbf{E}}^{k}_S-\text{Pool}^{m}(\bar{\textbf{E}}^{k}_S),
\end{equation}
\begin{equation}\label{7}
\textbf{E}^{k}_S = [\bar{\textbf{E}}^{k}_S, \bar{\textbf{E}}^{k,3}_{SC},..., \bar{\textbf{E}}^{k,9}_{SC}],
\end{equation}
\begin{equation}\label{8}
\bar{\textbf{E}}^{k}_D = \text{ReLU}(\text{BN}(\text{Conv}(\textbf{E}^{k}_{T1}-\textbf{E}^{k}_{T2}))),
\end{equation}
\begin{equation}\label{9}
\bar{\textbf{E}}^{k,m}_{DC} = \bar{\textbf{E}}^{k}_D-\text{Pool}^{m}(\bar{\textbf{E}}^{k}_D),
\end{equation}
\begin{equation}\label{10}
\textbf{E}^{k}_D = [\bar{\textbf{E}}^{k}_D, \bar{\textbf{E}}^{k,3}_{DC},..., \bar{\textbf{E}}^{k,9}_{DC}],
\end{equation}
where $\textbf{E}^{k}_S$ and $\textbf{E}^{k}_D$ $(k=1,2,...,5)$ are the enhanced features with point-wise summation and difference, respectively.
$\text{ReLU}$ is the rectified linear unit, $\text{BN}$ is the batch normalization, $\text{Conv}$ is a $1\times1$ convolution, and $\text{Pool}^{m}$ is a $m\times m$ average pooling with appropriate paddings $(m \in\{3, 5, 7, 9\})$.
[,] is the feature concatenation in channel.
In fact, $\bar{\textbf{E}}^{k,m}_{SC}$ and $\bar{\textbf{E}}^{k,m}_{DC}$ capture contrast features, and can make change regions stand out from their surrounding background.
Meanwhile, in most cases, keeping the original features shows better results due to the rich contextual information.
Thus, we concatenate them with contrast features.
Through the proposed DFE, more information of change regions and boundaries are highlighted with temporal information.
Thus, the framework can make the extracted features more discriminative and obtain better CD results.
We refer the readers to~\cite{luo2017non} for more insights.
\subsection{Progressive Change Prediction}
\begin{figure}
\centering
\resizebox{0.5\textwidth}{!}
{
\begin{tabular}{@{}c@{}c@{}}
\includegraphics[width=1\linewidth,height=0.46\linewidth]{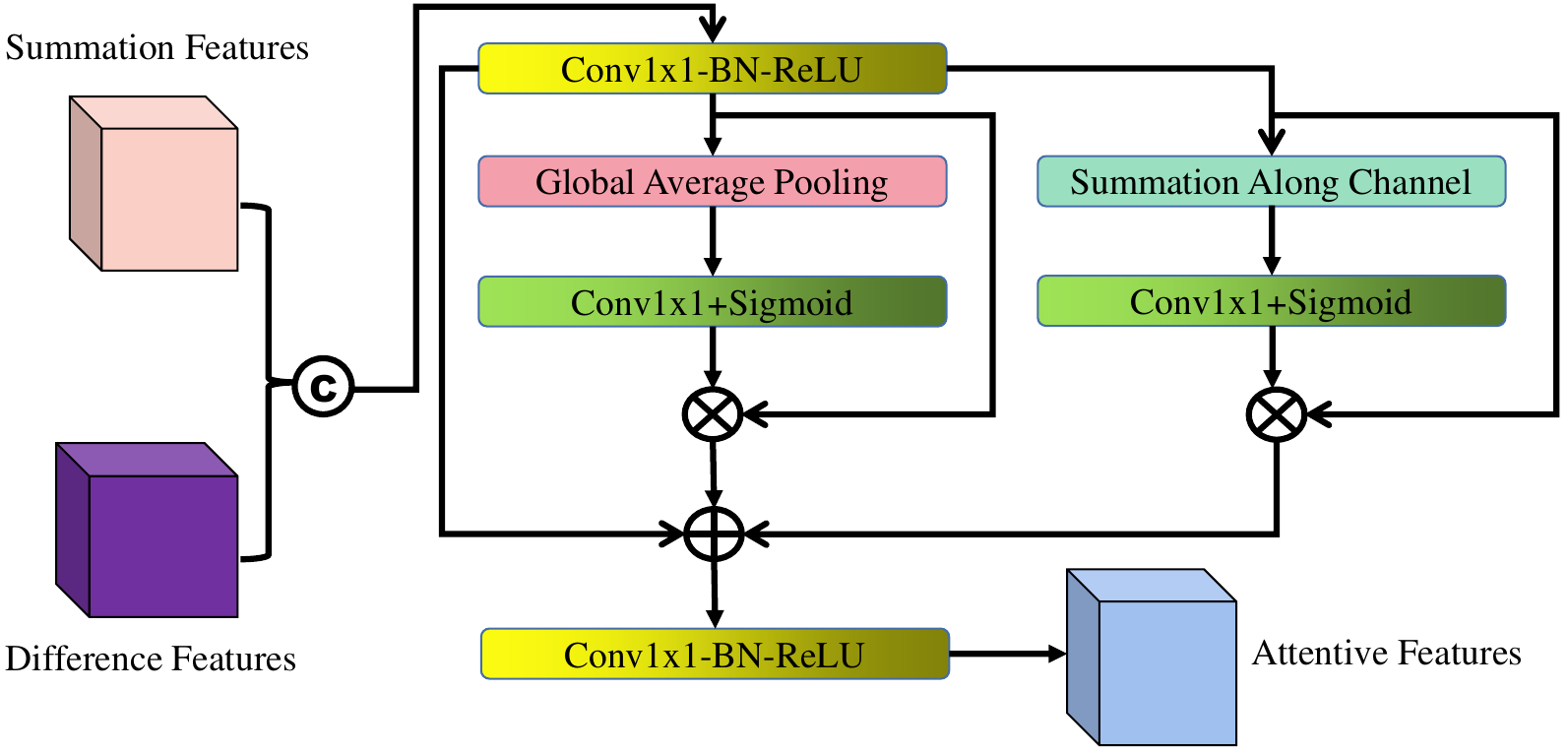} \\
\end{tabular}
}
\caption{The structure of our proposed Progressive Attention Module (PAM).}
\label{fig:PAM}
\end{figure}

Since change regions can be any shapes and appear at any scales, we should consider the CD predictions in various cases.
Inspired by the feature pyramid~\cite{lin2017feature}, we propose a progressive change prediction method, as shown in the middle part of Fig.~\ref{fig:Framework}.
To improve the representation ability, a pyramid structure with a Progressive Attention Module (PAM) is utilized with additional interdependencies through spatial and channel attentions.
The structure of the proposed PAM is illustrated in Fig.~\ref{fig:PAM}.
The PAM first takes the summation features and difference features as inputs, then a Spatial-level Attention (SA) and a Channel-level Attention (CA) are jointly applied to enhance the features related to change regions.
In addition, as we all know, residual connections~\cite{he2016deep} are famous, popular and efficient structures in current deep models. It alleviates the vanishing gradient problem and accelerates the training convergence. Thus, we further introduce a residual connection to improve the learning ability.
The final feature map can be obtained by a $1\times1$ convolution.
Formally, the PAM can be represented as:
\begin{equation}\label{11}
\textbf{F}^{k} = \text{ReLU}(\text{BN}(\text{Conv}([\textbf{E}^{k}_S, \textbf{E}^{k}_D]))),
\end{equation}
\begin{equation}\label{12}
\textbf{F}^{k}_{SA} = \textbf{F}^{k}*\sigma(\text{Conv}(\text{SAC}(\textbf{F}^{k}))),
\end{equation}
\begin{equation}\label{13}
\textbf{F}^{k}_{CA} = \textbf{F}^{k}*\sigma(\text{Conv}(\text{GAP}(\textbf{F}^{k}))),
\end{equation}
\begin{equation}\label{14}
\textbf{F}^{k}_A = \text{Conv}(\textbf{F}^{k}_{SA} +\textbf{F}^{k}_{CA} +\textbf{F}^{k}),
\end{equation}
where $\sigma$ is the Sigmoid function, $\text{SAC}$ is the summation along the channel, and $\text{GAP}$ is the global average pooling.

To achieve the progressive change prediction, we build the decoder pyramid grafted with a PAM as follows:
\begin{equation}\label{15}
\textbf{F}^{k}_P = \left\{
\begin{aligned}
&\textbf{F}^{k}_A,& \quad k=5,\\
&\text{UM}(\text{SwinBlock}^{n}(\textbf{F}^{k+1}_P)))+\textbf{F}^{k}_A,& \quad 1\le k<5.
\end{aligned}
\right.
\end{equation}
where $\text{UM}$ is the patch unmerging block for feature upsampling~\cite{liu2021swin}, and $\text{SwinBlock}^{n}$ is the Swin Transformer block with $n$ layers.
From the above formula, one can see that our PCP can make full use of the interdependencies within spatial and channel, and progressively aggregate multi-level features to improve the perception ability of the change regions.
Here, we note that the pyramid structure is complementary to residual connections.
Many previous works have already verified this fact. Thus, it is reasonable to introduce residual connections under the premise of a pyramid structure.
\subsection{Loss Function}
To optimize our framework, DS is utilized to achieve the deeply-supervised learning~\cite{zhang2017amulet,zhang2018agile,zhang2020non} with multiple boundary-aware loss functions for each feature level.
The overall loss is defined as the summation loss over all the side-outputs and the final fusion prediction.
Specifically, we first take the features of the PCP, \emph{i.e.}, $\textbf{F}^{k}_P$($k=1,2,...,5$), and use a deconvolutional layer for the corresponding prediction $\textbf{P}^{s}$ as side-outputs.
Then, we concatenate them for the final fusion prediction,
\begin{equation}\label{16}
\textbf{P}^{f} =\text{Conv}[\textbf{P}^{1},...,\textbf{P}^{S}].
\end{equation}
All the side-outputs and the final fusion prediction are supervised by the proposed hybrid loss:
\begin{equation}\label{17}
\mathcal{L} = \mathcal{L}^{f}+\sum^{S}_{s=1} \alpha_{s}\mathcal{L}^{s},
\end{equation}
where $\mathcal{L}^{f}$ is the loss of the final fusion prediction and $\mathcal{L}^{s}$ is the loss of the $s$-th side-output, respectively.
$S$ denotes the total number of the side-outputs and $\alpha_{s}$ is the weight for each level loss.
Our method includes five side-outputs, \emph{i.e.}, $S=5$.

To obtain complete CD regions and regular CD boundaries, we define $\mathcal{L}^{f}$ or $\mathcal{L}^{s}$ as a combined loss with three terms:
\begin{equation}\label{18}
\mathcal{L}^{f/s} = \mathcal{L}_{WBCE}+\mathcal{L}_{SSIM}+\mathcal{L}_{SIoU},
\end{equation}
where $\mathcal{L}_{WBCE}$ is the weighted binary cross-entropy loss, $\mathcal{L}_{SSIM}$ is the structural similarity loss and $\mathcal{L}_{SIoU}$ is the soft intersection over union loss.
The $\mathcal{L}_{WBCE}$ provides a probabilistic measure of similarity between the prediction and ground truth from a pixel-level view.
The $\mathcal{L}_{SSIM}$ captures the structural information of change regions in patch-level.
The $\mathcal{L}_{SIoU}$ is inspired by measuring the similarity of two sets, and yields a global similarity in CD map-level.
More specifically, given the ground truth probability $g_{l}(\textbf{x})$ and the estimated probability $p_{l}(\textbf{x})$ at pixel $\textbf{x}$ belong to the class $l$, the $\mathcal{L}_{WBCE}$ loss function is defined as:
\begin{equation}\label{19}
\mathcal{L}_{WBCE}=-\sum_{\textbf{x}}w(\textbf{x})g_{l}(\textbf{x})\text{log}(p_{l}(\textbf{x})).
\end{equation}
Here, we utilize weights $w(\textbf{x})$ to adapt the loss function to the challenges that we have encountered in CD: the class imbalance and the errors along CD boundaries.
Given the frequency $f_l$ of class $l$ in the training data, the indicator function $I$, the training prediction $T$, and the gradient operator $\nabla$, then the weights are defined as:
\begin{equation}\label{20}
w(\textbf{x})=\sum_{l}I(T(\textbf{x}==l))\frac{median(\textbf{f})}{f_l}+w_{0}I(|\nabla T(\textbf{x})|>0),
\end{equation}
where $\textbf{f} = [f_1,..., f_L]$ is the vector of all class frequencies.
The first term models the median frequency balancing~\cite{badrinarayanan2017segnet} and compensates for the class imbalance problem by highlighting classes with a low probability.
The second term puts higher weights on the CD boundaries to emphasize on the correct detection of contours.

The $\mathcal{L}_{SSIM}$ loss considers a local neighborhood of each pixel~\cite{wang2003multiscale}.
Let $\hat{\textbf{x}} = \{\textbf{x}_j: j = 1, ...,N^2\}$ and $\hat{\textbf{y}} = \{\textbf{y}_j: j = 1, ...,N^2\}$
be the pixel values of two corresponding patches (size: $N\times N$) cropped from the prediction $P$ and the ground truth $G$ respectively, the $\mathcal{L}_{SSIM}$ loss is defined as:
\begin{equation}\label{21}
\mathcal{L}_{SSIM}=1-\frac{(2\mu_{\textbf{x}}\mu_{\textbf{x}}+\epsilon)(2\sigma_{\textbf{xy}}+\epsilon)}{(\mu^{2}_{\textbf{x}}+\mu^{2}_{\textbf{y}}+\epsilon)(\sigma^{2}_{\textbf{x}}+\sigma^{2}_{\textbf{y}}+\epsilon)},
\end{equation}
where $\mu_{\textbf{x}}$, $\mu_{\textbf{y}}$ and $\sigma_{\textbf{x}}$, $\sigma_{\textbf{y}}$ are the mean and standard deviations
of $\hat{\textbf{x}}$ and $\hat{\textbf{y}}$ respectively.
$\sigma_{\textbf{xy}}$ is their covariance.
Here, $\epsilon=10^{-4}$ is used to avoid dividing by zero.

In this work, one metric of interest at test time is the Intersection over Union (IoU).
Thus, we also introduce the soft IoU loss~\cite{mattyus2017deeproadmapper}, which is differentiable for model learning.
The $\mathcal{L}_{SIoU}$ is defined as:
\begin{equation}\label{22}
\mathcal{L}_{SIoU}=1-\frac{\sum_{\textbf{x}}p_{l}(\textbf{x})g_{l}(\textbf{x})}{\sum_{\textbf{x}}[p_{l}(\textbf{x})+g_{l}(\textbf{x})-p_{l}(\textbf{x})g_{l}(\textbf{x})]}.
\end{equation}
When utilizing all above losses, the $\mathcal{L}_{WBCE}$ loss can relieve the class imbalance problem for change pixels, the $\mathcal{L}_{SSIM}$ loss highlights the local structure of change boundaries, and the $\mathcal{L}_{SIoU}$ loss gives more focus on the overall change regions.
Thus, we can obtain better CD results and make the framework easier to optimize.
\section{Experiments}
In this section, we perform extensive experiments to verify the effectiveness of the proposed framework.
We first introduce the used datasets, evaluation metrics and implementation details.
Then, we compare the proposed method with other outstanding CD methods.
Finally, we perform ablation studies to verify the effectiveness of key modules with quantitative and qualitative comparisons.
\subsection{Datasets}
\textbf{LEVIR-CD}~\cite{chen2020spatial} is a public large-scale remote sensing CD dataset.
It contains 637 image pairs with a 1024$\times$1024 resolution (0.5m).
We follow its default dataset split, and crop original images into small patches of size 256$\times$256 with no overlapping.
Therefore, we obtain 7120/1024/2048 pairs of image patches for training/validation/test, respectively.

\textbf{WHU-CD}~\cite{ji2018fully} is a public building CD dataset.
It contains one pair of high-resolution (0.075m) aerial images of size 32507$\times$15354.
As no pre-definite data split is widely-used, we crop the original image into small patches of size 256$\times$256 with
no overlap and randomly split them into three parts: 6096/762/762 for training/validation/test, respectively.

\textbf{SYSU-CD}~\cite{shi2021deeply} is also a public building CD dataset.
It contains 20000 pairs of high-resolution (0.5m) images of size 256$\times$256.
We follow its default dataset split for experiments.
There are 12000/4000/4000 pairs of image patches for training/validation/test, respectively.

\textbf{Google-CD}~\cite{liu2021super} is a very recent and public CD dataset.
It contains 19 image pairs, originating from Google Earth Map.
The image resolutions range from 1006$\times$1168 pixels to 4936$\times$5224 pixels.
As WHU-CD, we also crop the original images into small patches of size 256$\times$256 with no overlap and randomly split them into three parts: 2504/313/313
for training/validation/test, respectively.

In addition, we adopt two SAR image CD datasets~\cite{gao2019sea} to verify the generalization of our method.
These two datasets were gathered from disaster-stricken environments, which are related to flood and ice breakup situations, respectively.
The Ottawa dataset is captured by the RADARSAT SAR sensor in May and August 1997 and changes are caused by floods.
The size of each image is 290$\times$350 pixels.
The Sulzberger dataset is a part of Sulzberger Ice Shelf, which is provided by the European Space Agency’s Envisat satellite.
The size of the image is 256$\times$256 pixels.
It clearly shows the breakup of an ice shelf caused by a tsunami in March 2011.
\subsection{Evaluation Metrics}
To verify the performance of our framework and other compared methods, we follow previous works~\cite{chen2021remote,ke2022hybrid,wang2022network,bandara2022transformer} and utilize F1 and Intersection over Union (IoU) scores with regard to the change-class as
the primary evaluation metrics.
Additionally, we also report the precision and recall of the change category, Overall Accuracy (OA) and Receiver Operating Characteristic (ROC) curve.
To evaluate the performance of regional boundaries, we follow previous works~\cite{cheng2020cascadepsp,shen2022high} and adopt the mean Boundary Accuracy (mBA) as the metric.
\begin{table*}
\centering
\caption{Quantitative comparisons on LEVIR-CD and WHU-CD datasets. The best and the second best are in bold and underline, respectively. $-$ means the results of corresponding methods are missing.}
\label{tab:LEVIR}
\resizebox{0.84\textwidth}{!}
{
\centering
 \begin{tabular}{l|c|c|c|c|c|c|c|c|c|ccccccccc}
 \hline
\multirow{3}{*}{Methods} & \multicolumn{5}{c|}{LEVIR-CD}&\multicolumn{5}{c}{WHU-CD}\\
\cline{2-11}
~&Pre.&Rec.&F1&IoU&OA&Pre.&Rec.&F1&IoU&OA\\
\hline
FC-EF~\cite{daudt2018fully}        &86.91&80.17&83.40&71.53&98.39&71.63&67.25&69.37&53.11&97.61\\
FC-Siam-Diff~\cite{daudt2018fully}   &89.53&83.31&86.31&75.92&98.67&47.33&77.66&58.81&41.66&95.63\\
FC-Siam-Conc~\cite{daudt2018fully}   &91.99&76.77&83.69&71.96&98.49&60.88&73.58&66.63&49.95&97.04\\
BiDateNet~\cite{liu2020building}   &85.65&89.98&87.76&78.19&98.52&78.28&71.59&74.79&59.73&81.92\\
U-Net++MSOF~\cite{peng2019end}   &90.33&81.82&85.86&75.24&98.41&91.96&89.40&90.66&82.92&96.98\\
DTCDSCN~\cite{liu2020building}   &88.53&86.83&87.67&78.05&98.77&63.92&82.30&71.95&56.19&97.42\\
DASNet~\cite{liu2020building}   &80.76&79.53&79.91&74.65&94.32&68.14&73.03&70.50&54.41&97.29\\
STANet~\cite{chen2020spatial}   &83.81&\textbf{91.00}&87.26&77.40&98.66&79.37&85.50&82.32&69.95&98.52\\
MSTDSNet~\cite{song2022mstdsnet}   &85.52&\underline{90.84}&88.10&78.73&98.56&-----&-----&-----&-----&-----\\
IFNet~\cite{zhang2020deeply}   &\textbf{94.02}&82.93&88.13&78.77&98.87&\textbf{96.91}&73.19&83.40&71.52&98.83\\
SNUNet~\cite{fang2021snunet}  &89.18&87.17&88.16&78.83&98.82&85.60&81.49&83.50&71.67&98.71\\
\hline
BIT~\cite{chen2021remote}   &89.24&89.37&89.31&80.68&98.92&86.64&81.48&83.98&72.39&98.75\\
H-TransCD~\cite{ke2022hybrid}
&91.45&88.72&90.06&81.92&99.00&93.85&88.73&91.22&83.85&99.24\\
UVACD~\cite{wang2022network} &91.90&90.70&\underline{91.30}&\textbf{83.98}&\textbf{99.12}&91.45&88.72&90.06&81.92&99.00\\
ChangeFormer~\cite{bandara2022transformer}   &92.05&88.80&90.40&82.48&99.04&91.83&88.02&89.88&81.63&99.12\\
FTN~\cite{yan2022fully}
&92.71&89.37&91.01&83.51&99.06&93.09&\underline{91.24}&\underline{92.16}&\underline{85.45}&\underline{99.37}\\
Ours
&\underline{92.90}&89.35&\textbf{91.90}&\underline{83.64}&\underline{99.07}&\underline{94.68}&\textbf{92.12}&\textbf{93.38}&\textbf{87.58}&\textbf{99.47}\\
\hline
\end{tabular}
}
\end{table*}
\begin{table*}
\centering
\caption{Quantitative comparisons on SYSU-CD and Google-CD datasets. The best and the second best are in bold and underline, respectively. $-$ means the results of corresponding methods are missing.}
\label{tab:SYSU}
\resizebox{0.84\textwidth}{!}
{
\centering
 \begin{tabular}{l|c|c|c|c|c|c|c|c|c|ccccccccccc}
 \hline
\multirow{3}{*}{Methods} & \multicolumn{5}{c|}{SYSU-CD}&\multicolumn{5}{c}{Google-CD}\\
\cline{2-11}
~&Pre.&Rec.&F1&IoU&OA&Pre.&Rec.&F1&IoU&OA\\
\hline
FC-EF~\cite{daudt2018fully}        &74.32&75.84&75.07&60.09&86.02&80.81&64.39&71.67&55.85&85.85\\
FC-Siam-Diff~\cite{daudt2018fully}   &\textbf{89.13}&61.21&72.57&56.96&82.11&85.44&63.28&72.71&57.12&87.27\\
FC-Siam-Conc~\cite{daudt2018fully}   &82.54&71.03&76.35&61.75&86.17&82.07&64.73&72.38&56.71&84.56\\
BiDateNet~\cite{liu2020building}   &81.84&72.60&76.94&62.52&89.74&78.28&71.59&74.79&59.73&81.92\\
U-Net++MSOF~\cite{peng2019end}   &81.36&75.39&78.26&62.14&86.39&\underline{91.21}&57.60&70.61&54.57&95.21\\
DASNet~\cite{liu2020building}   &68.14&70.01&69.14&60.65&80.14&71.01&44.85&54.98&37.91&90.87\\
STANet~\cite{chen2020spatial}   &70.76&\textbf{85.33}&77.37&63.09&87.96&89.37&65.02&75.27&60.35&82.58\\
DSAMNet~\cite{zhang2020deeply}   &74.81&\underline{81.86}&78.18&64.18&89.22&72.12&80.37&76.02&61.32&94.93\\
MSTDSNet~\cite{song2022mstdsnet}   &79.91&80.76&80.33&67.13&90.67&-----&-----&-----&-----&-----\\
SRCDNet~\cite{liu2021super}   &75.54&81.06&78.20&64.21&89.34&83.74&71.49&77.13&62.77&83.18\\
\hline
BIT~\cite{chen2021remote}   &82.18&74.49&78.15&64.13&90.18&\textbf{92.04}&72.03&80.82&67.81&96.59\\
H-TransCD~\cite{ke2022hybrid}
&83.05&77.40&80.13&66.84&90.95&85.93&81.73&83.78&72.08&97.64\\
FTN~\cite{yan2022fully}
&86.86&76.82&\underline{81.53}&\underline{68.82}&\underline{91.79}&86.99&\textbf{84.21}&\underline{85.58}&\underline{74.79}&\underline{97.92}\\
Ours
&\underline{89.09}&77.42&\textbf{82.84}&\textbf{70.71}&\textbf{92.44}&87.98&\underline{84.18}&\textbf{86.04}&\textbf{75.50}&\textbf{97.97}\\
\hline
\end{tabular}
}
\end{table*}
\subsection{Implementation Details}
We perform experiments with the PyTorch toolbox and one NVIDIA A30 GPU.
We use the mini-batch SGD algorithm to train our framework with an initial learning rate $10^{-3}$, moment 0.9 and weight decay 0.0005.
The batch size is set to 6.
For the Siamese feature extraction backbone, we adopt the Swin Transformer pre-trained on ImageNet-22k classification task~\cite{deng2009imagenet}.
To fit the input size of the pre-trained Swin Transformer, we uniformly resize image patches to 384$\times$384.
For other layers, we randomly initialize them and set the learning rate with 10 times than the initial learning rate.
We train the framework with 100 epochs.
The learning rate decreases to the 1/10 of the initial learning rate at every 20 epoch.
To improve the robustness, data augmentation is performed by random rotation and flipping of the input images.
For the loss function in the model training, the weight parameters of each level are set equally.
We release the source code at https://github.com/Drchip61/TransYNet.
\subsection{Comparisons with State-of-the-arts}
In this section, we compare the proposed method with other outstanding methods four optical and two SAR image CD datasets.
These experimental results fully verify the effectiveness of our proposed framework and modules.

\begin{figure}
\centering
\resizebox{0.46\textwidth}{!}
{
\begin{tabular}{@{}c@{}c@{}}
\includegraphics[width=1\linewidth,height=0.64\linewidth]{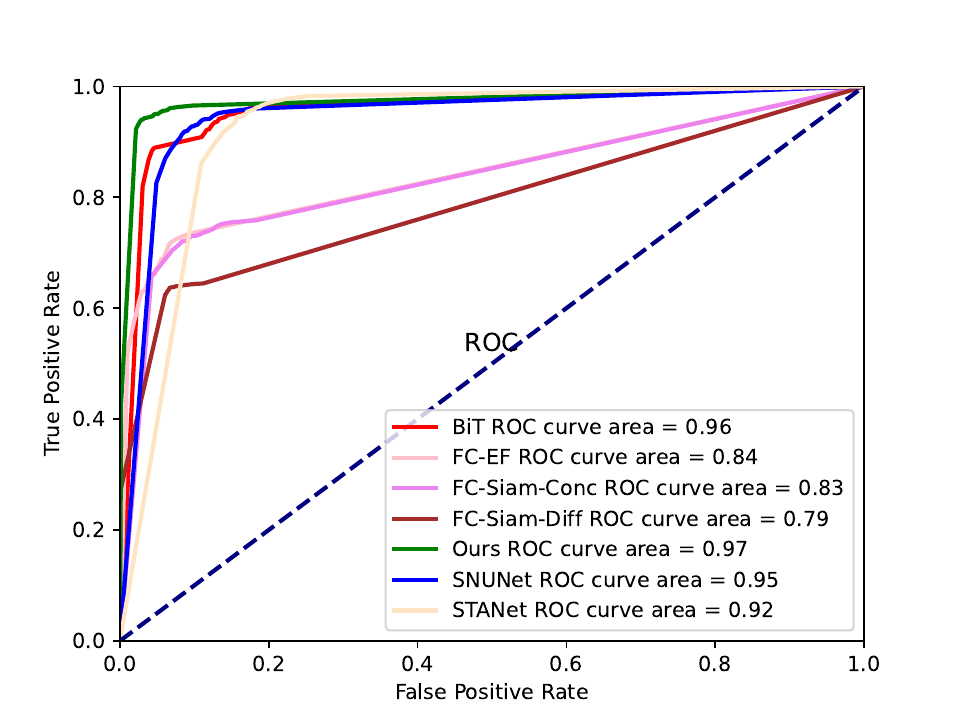} \\
\end{tabular}
}
\caption{The ROC curves of some typical methods on the WHU-CD dataset.}
\label{fig:roc}
\end{figure}

\textbf{Quantitative Comparisons.}
We present the comparative results in Tab.~\ref{tab:LEVIR} and Tab.~\ref{tab:SYSU}.
The results clearly show that our method delivers excellent performance.
More specifically, our method achieves the F1 and IoU scores of 91.90\% and 83.64\% on the LEVIR-CD dataset, respectively.
They are much better than most of previous methods.
Besides, compared with other Transformer-based methods, such as BIT~\cite{chen2021remote}, H-TransCD~\cite{ke2022hybrid} and ChangeFormer~\cite{bandara2022transformer}, our method shows consistent improvements in terms of all evaluation metrics.
When compared with our previous method FTN~\cite{yan2022fully}, the method in this paper can achieve better results in almost all metrics.
On the WHU-CD dataset, our method shows significant improvements with the F1 and IoU scores of 93.38\% and 87.58\%, respectively.
In comparison with the second-best method (FTN), our method improves the F1 and IoU scores by 1.2\% and 2.1\%, respectively.
On the SYSU-CD dataset, our method achieves the F1 and IoU scores of 82.84\% and 70.71\%, respectively.
The SYSU-CD dataset includes more large-scale change regions.
We believe that the improvements are mainly based on the proposed DFE.
On the Google-CD dataset, our method shows much better results than other compared methods.
In fact, our method achieves the F1 and IoU scores of 86.04\% and 75.50\%, respectively.
We note that the Google-CD dataset is recently proposed and it is much challenging than other three datasets.
We also note that the performance of precision, recall and OA is not consistent in all methods.
Our method generally achieves better recall values than most compared methods.
The main reason may be that our method gives higher confidences to the change regions.
To better illustrate the performance, we also present the ROC curves of some typical methods in Fig.~\ref{fig:roc}.
It is observed that our method achieves better results than other typical methods on the WHU-CD dataset.

\begin{figure*}
\centering
\resizebox{1\textwidth}{!}
{
\begin{tabular}{@{}c@{}c@{}c@{}c@{}c@{}c@{}c@{}c@{}c@{}c@{}c@{}c}
\vspace{-2mm}
\includegraphics[width=0.1\linewidth,height=1.6cm]{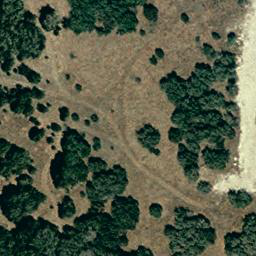}\ &
\includegraphics[width=0.1\linewidth,height=1.6cm]{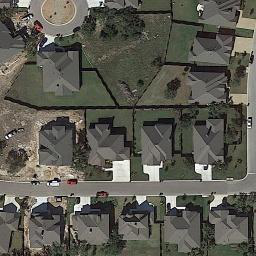}\ &
\includegraphics[width=0.1\linewidth,height=1.6cm]{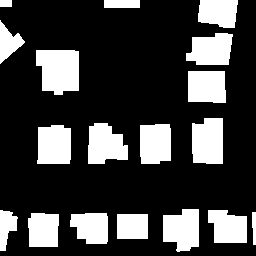}\ &
\includegraphics[width=0.1\linewidth,height=1.6cm]{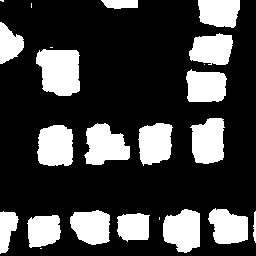}\ &
\includegraphics[width=0.1\linewidth,height=1.6cm]{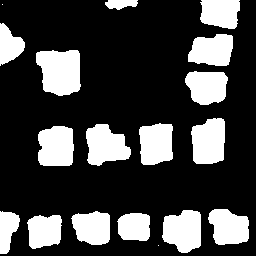}\ &
\includegraphics[width=0.1\linewidth,height=1.6cm]{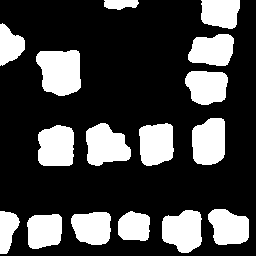}\ &
\includegraphics[width=0.1\linewidth,height=1.6cm]{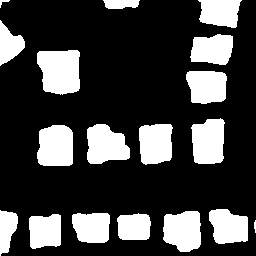}\ &
\includegraphics[width=0.1\linewidth,height=1.6cm]{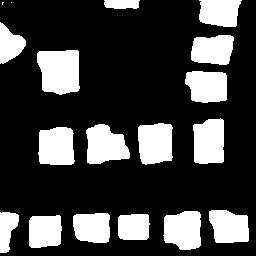}\ \\
\vspace{-2mm}
\includegraphics[width=0.1\linewidth,height=1.6cm]{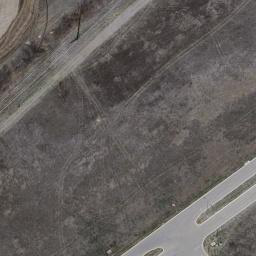}\ &
\includegraphics[width=0.1\linewidth,height=1.6cm]{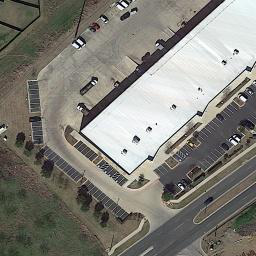}\ &
\includegraphics[width=0.1\linewidth,height=1.6cm]{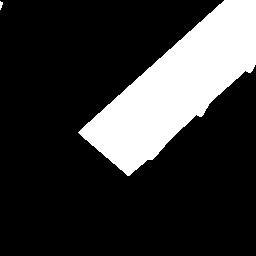}\ &
\includegraphics[width=0.1\linewidth,height=1.6cm]{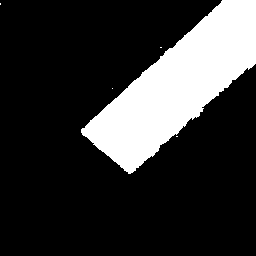}\ &
\includegraphics[width=0.1\linewidth,height=1.6cm]{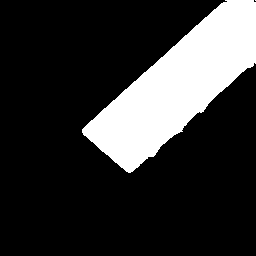}\ &
\includegraphics[width=0.1\linewidth,height=1.6cm]{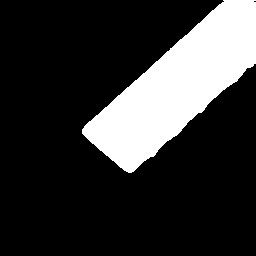}\ &
\includegraphics[width=0.1\linewidth,height=1.6cm]{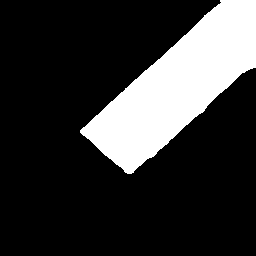}\ &
\includegraphics[width=0.1\linewidth,height=1.6cm]{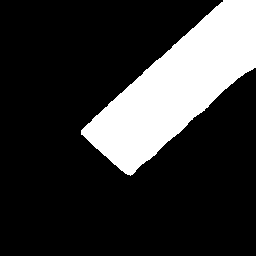}\ \\
 {\tiny T1 Image} & {\tiny T2 Image} & {\tiny Ground Truth} & {\tiny Ours} & {\tiny FC-Conc} & {\tiny FC-Diff} & {\tiny DTCDSCN} & {\tiny BIT}\ \\
\end{tabular}
}
\caption{Comparison of typical change detection results on the LEVIR-CD dataset. Best view by zooming in.}
\label{fig:LEVIR-CD}
\end{figure*}

\begin{figure*}
\centering
\resizebox{1\textwidth}{!}
{
\begin{tabular}{@{}c@{}c@{}c@{}c@{}c@{}c@{}c@{}c@{}c@{}c@{}c@{}c}
\vspace{-2mm}
\includegraphics[width=0.1\linewidth,height=1.6cm]{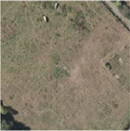}\ &
\includegraphics[width=0.1\linewidth,height=1.6cm]{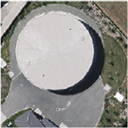}\ &
\includegraphics[width=0.1\linewidth,height=1.6cm]{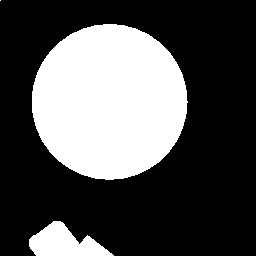}\ &
\includegraphics[width=0.1\linewidth,height=1.6cm]{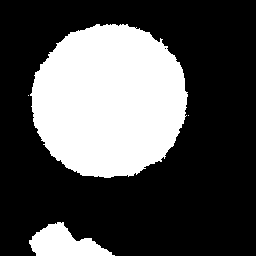}\ &
\includegraphics[width=0.1\linewidth,height=1.6cm]{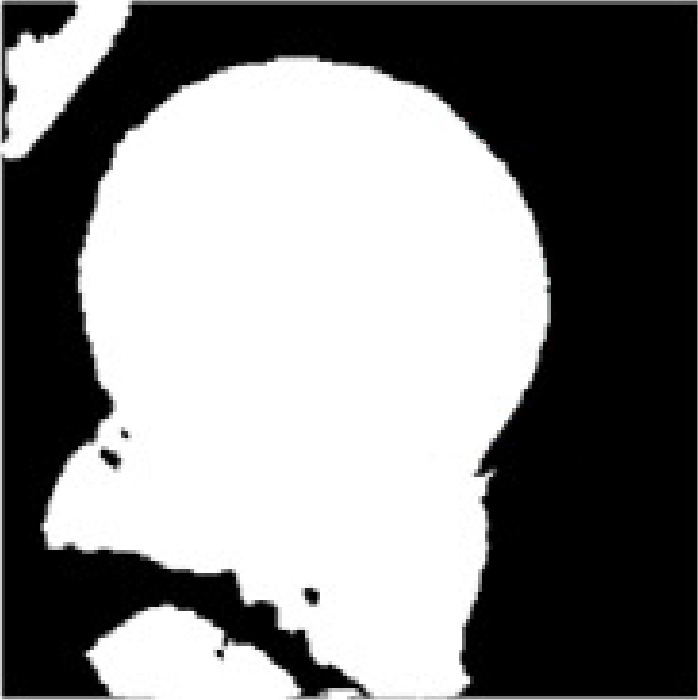}\ &
\includegraphics[width=0.1\linewidth,height=1.6cm]{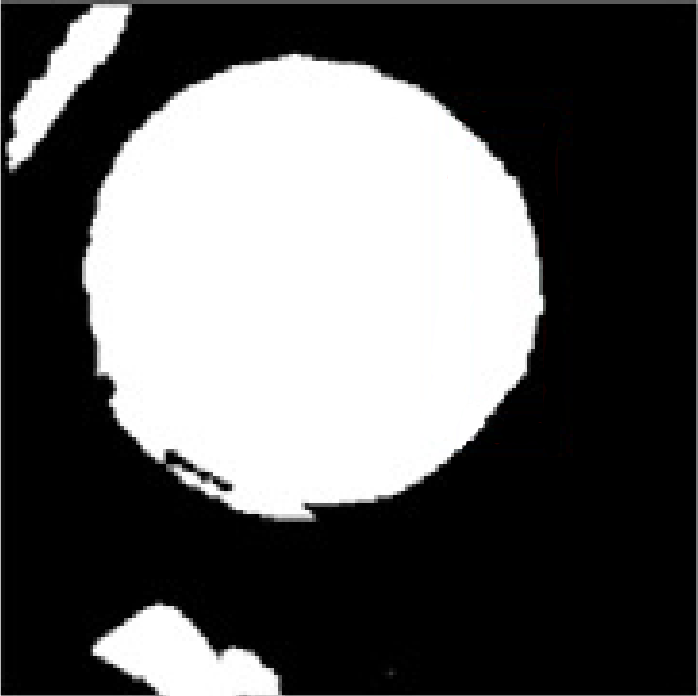}\ &
\includegraphics[width=0.1\linewidth,height=1.6cm]{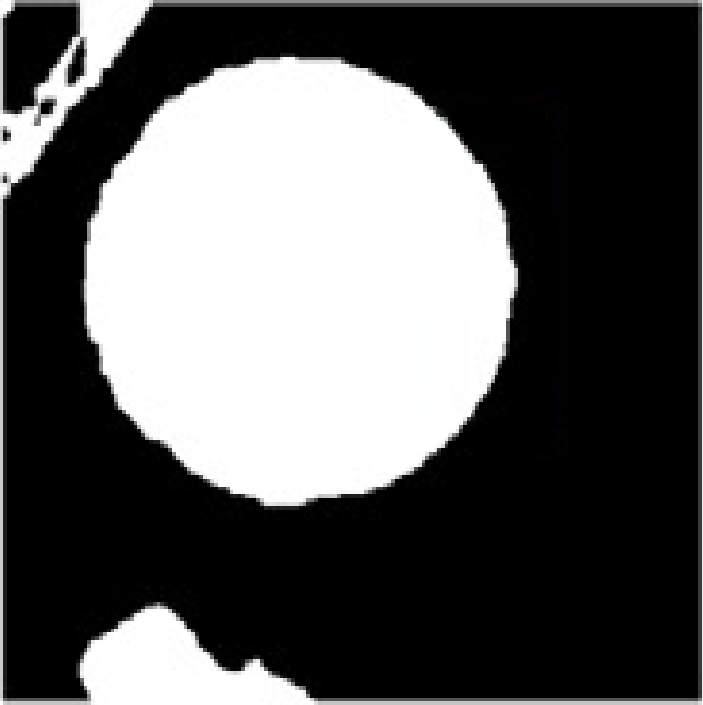}\ &
\includegraphics[width=0.1\linewidth,height=1.6cm]{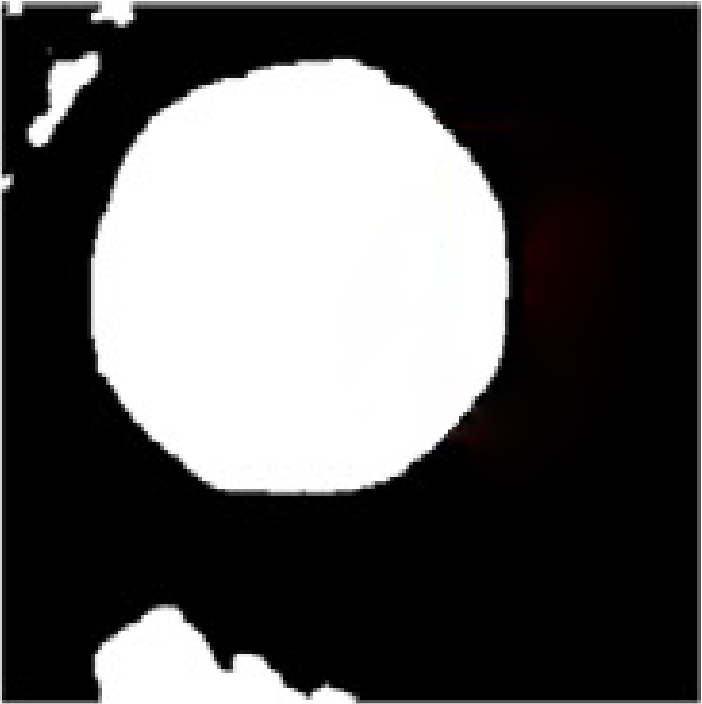}\ \\
\vspace{-2mm}
\includegraphics[width=0.1\linewidth,height=1.6cm]{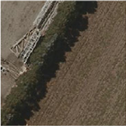}\ &
\includegraphics[width=0.1\linewidth,height=1.6cm]{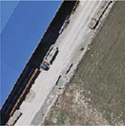}\ &
\includegraphics[width=0.1\linewidth,height=1.6cm]{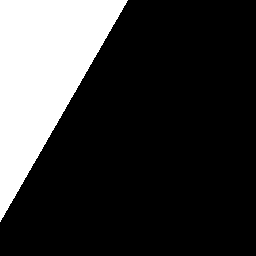}\ &
\includegraphics[width=0.1\linewidth,height=1.6cm]{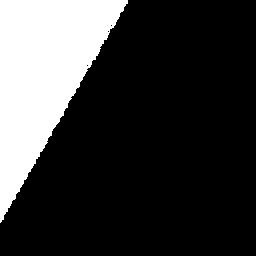}\ &
\includegraphics[width=0.1\linewidth,height=1.6cm]{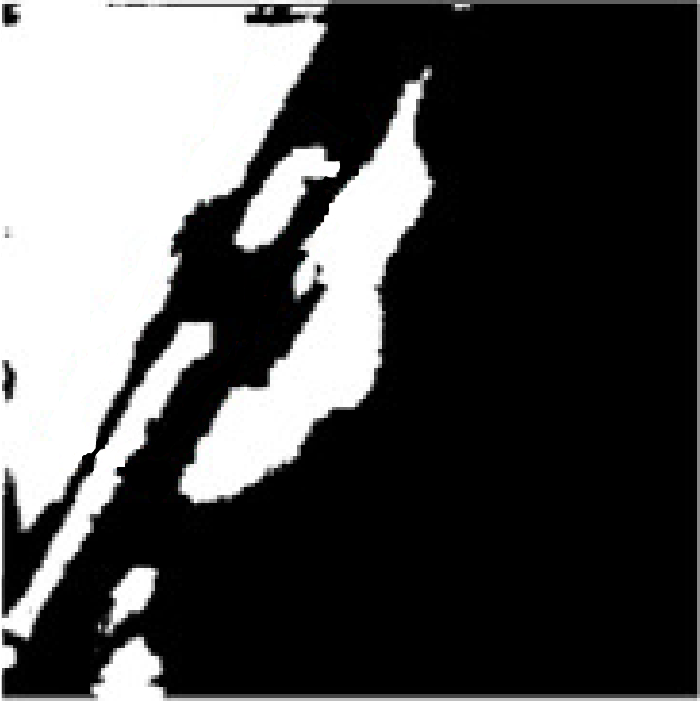}\ &
\includegraphics[width=0.1\linewidth,height=1.6cm]{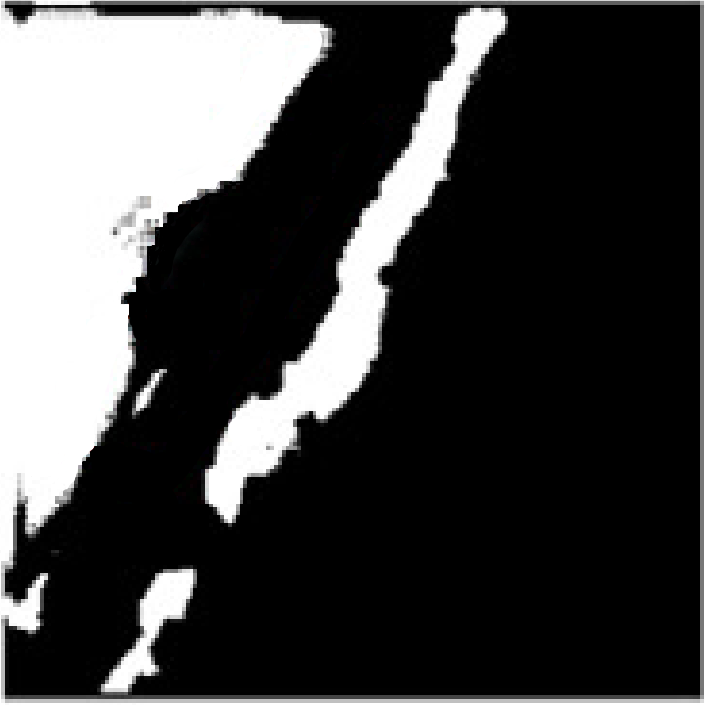}\ &
\includegraphics[width=0.1\linewidth,height=1.6cm]{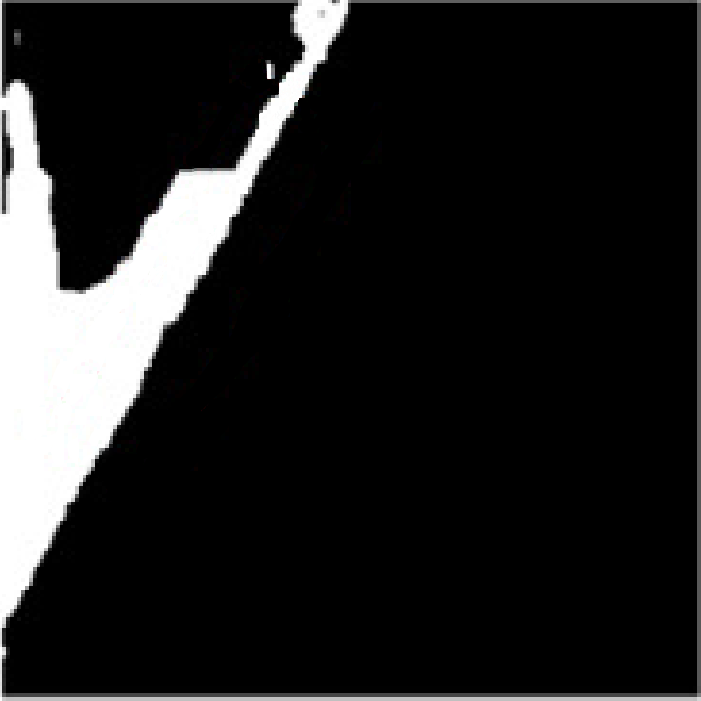}\ &
\includegraphics[width=0.1\linewidth,height=1.6cm]{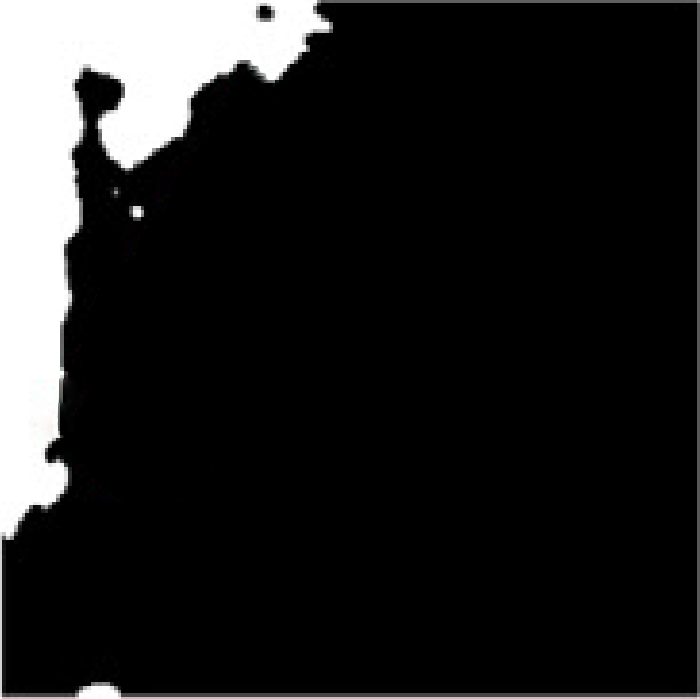}\ \\
 {\tiny T1 Image} & {\tiny T2 Image} & {\tiny Ground Truth} & {\tiny Ours} & {\tiny FC-Conc} & {\tiny FC-Diff} & {\tiny STANet} & {\tiny BIT}\ \\
\end{tabular}
}
\caption{Comparison of typical change detection results on the WHU-CD dataset. Best view by zooming in.}
\label{fig:WHU-CD}
\end{figure*}

\begin{figure*}
\centering
\resizebox{1\textwidth}{!}
{
\begin{tabular}{@{}c@{}c@{}c@{}c@{}c@{}c@{}c@{}c@{}c@{}c@{}c@{}c}
\vspace{-2mm}
\includegraphics[width=0.1\linewidth,height=1.6cm]{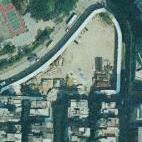}\ &
\includegraphics[width=0.1\linewidth,height=1.6cm]{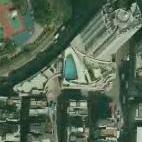}\ &
\includegraphics[width=0.1\linewidth,height=1.6cm]{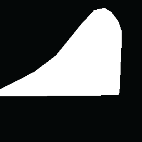}\ &
\includegraphics[width=0.1\linewidth,height=1.6cm]{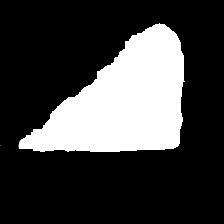}\ &
\includegraphics[width=0.1\linewidth,height=1.6cm]{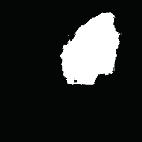}\ &
\includegraphics[width=0.1\linewidth,height=1.6cm]{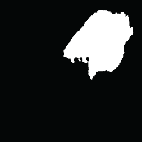}\ &
\includegraphics[width=0.1\linewidth,height=1.6cm]{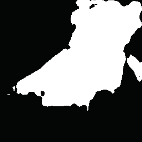}\ &
\includegraphics[width=0.1\linewidth,height=1.6cm]{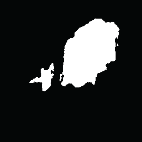}\ \\
\vspace{-2mm}
\includegraphics[width=0.1\linewidth,height=1.6cm]{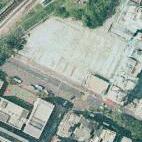}\ &
\includegraphics[width=0.1\linewidth,height=1.6cm]{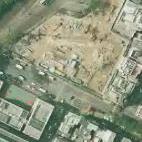}\ &
\includegraphics[width=0.1\linewidth,height=1.6cm]{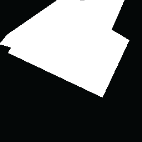}\ &
\includegraphics[width=0.1\linewidth,height=1.6cm]{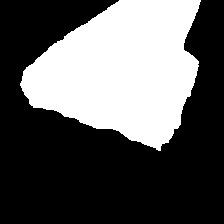}\ &
\includegraphics[width=0.1\linewidth,height=1.6cm]{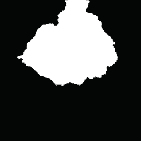}\ &
\includegraphics[width=0.1\linewidth,height=1.6cm]{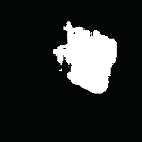}\ &
\includegraphics[width=0.1\linewidth,height=1.6cm]{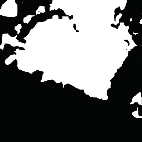}\ &
\includegraphics[width=0.1\linewidth,height=1.6cm]{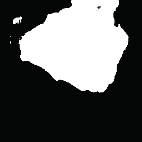}\ \\
 {\tiny T1 Image} & {\tiny T2 Image} & {\tiny Ground Truth} & {\tiny Ours} & {\tiny FC-Conc} & {\tiny FC-Diff} & {\tiny STANet} & {\tiny BiDateNet}\ \\
\end{tabular}
}
\caption{Comparison of typical change detection results on the SYSU-CD dataset. Best view by zooming in.}
\label{fig:SYSU-CD}
\end{figure*}

\begin{figure*}
\resizebox{1\textwidth}{!}
{
\begin{tabular}{@{}c@{}c@{}c@{}c@{}c@{}c@{}c@{}c@{}c@{}c@{}c@{}c}
\vspace{-2mm}
\includegraphics[width=0.1\linewidth,height=1.6cm]{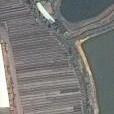}\ &
\includegraphics[width=0.1\linewidth,height=1.6cm]{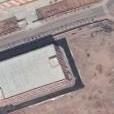}\ &
\includegraphics[width=0.1\linewidth,height=1.6cm]{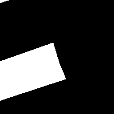}\ &
\includegraphics[width=0.1\linewidth,height=1.6cm]{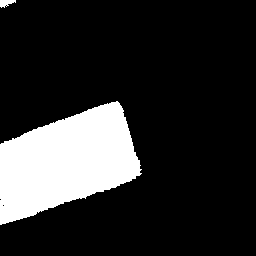}\ &
\includegraphics[width=0.1\linewidth,height=1.6cm]{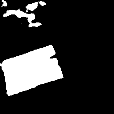}\ &
\includegraphics[width=0.1\linewidth,height=1.6cm]{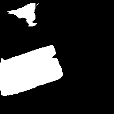}\ &
\includegraphics[width=0.1\linewidth,height=1.6cm]{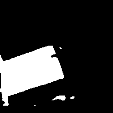}\ &
\includegraphics[width=0.1\linewidth,height=1.6cm]{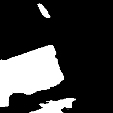}\ \\
\vspace{-2mm}
\includegraphics[width=0.1\linewidth,height=1.6cm]{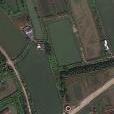}\ &
\includegraphics[width=0.1\linewidth,height=1.6cm]{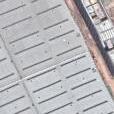}\ &
\includegraphics[width=0.1\linewidth,height=1.6cm]{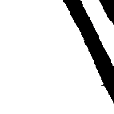}\ &
\includegraphics[width=0.1\linewidth,height=1.6cm]{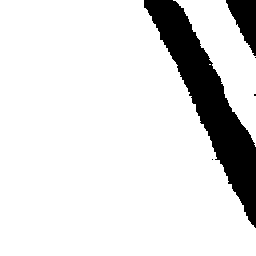}\ &
\includegraphics[width=0.1\linewidth,height=1.6cm]{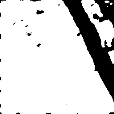}\ &
\includegraphics[width=0.1\linewidth,height=1.6cm]{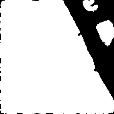}\ &
\includegraphics[width=0.1\linewidth,height=1.6cm]{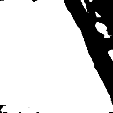}\ &
\includegraphics[width=0.1\linewidth,height=1.6cm]{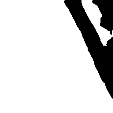}\ \\
 {\tiny T1 Image} & {\tiny T2 Image} & {\tiny Ground Truth} & {\tiny Ours} & {\tiny FC-Conc} & {\tiny FC-Diff} & {\tiny STANet} & {\tiny BiDateNet}\ \\
\end{tabular}
}
\caption{Comparison of typical change detection results on the Google-CD dataset. Best view by zooming in.}
\label{fig:Google-CD}
\end{figure*}

\textbf{Qualitative Comparisons.}
To illustrate the visual effect, we first display some typical CD results on the four optical image CD datasets, as shown in Fig.~\ref{fig:LEVIR-CD}-\ref{fig:Google-CD}.
From the results, one can see that our method generally shows best results.
For example, when change regions have multiple scales, our method can correctly identify most of the change regions, as shown in Fig.~\ref{fig:LEVIR-CD}.
When change objects cover most of the image regions, most of current methods can not detect them. However, our method can still detect them with clear boundaries, as shown in Fig.~\ref{fig:WHU-CD}.
In addition, when the change regions appear in complex scenes, our method can maintain the contour shape.
While most of compared methods fail, as shown in Fig.~\ref{fig:SYSU-CD}.
When distractors appear in the scene, our method can reduce the effect and correctly detect the real change regions, as shown in Fig.~\ref{fig:Google-CD}.
From the above visual results, we can see that our method shows superior performance than most methods.
\begin{figure*}
\centering
\resizebox{1\textwidth}{!}
{
\begin{tabular}{@{}c@{}c@{}c@{}c@{}c@{}c@{}c@{}c@{}c@{}c@{}c@{}c}
\vspace{-2mm}
\includegraphics[width=0.1\linewidth,height=1.2cm]{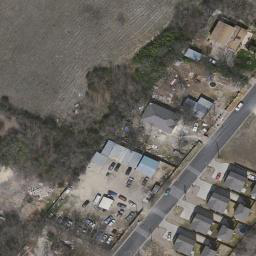}\ &
\includegraphics[width=0.1\linewidth,height=1.2cm]{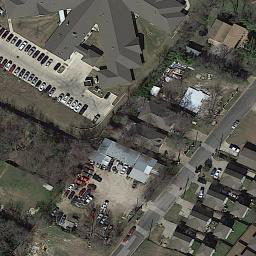}\ &
\includegraphics[width=0.1\linewidth,height=1.2cm]{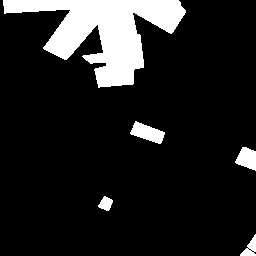}\ &
\includegraphics[width=0.1\linewidth,height=1.2cm]{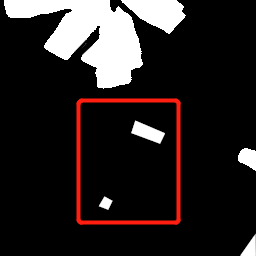}\ &
\includegraphics[width=0.1\linewidth,height=1.2cm]{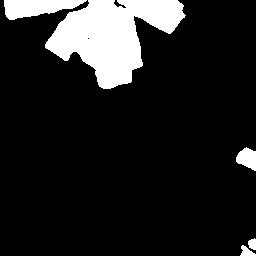}\ &
\includegraphics[width=0.1\linewidth,height=1.2cm]{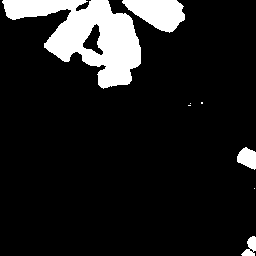}\ &
\includegraphics[width=0.1\linewidth,height=1.2cm]{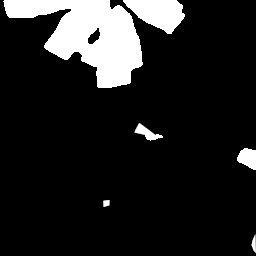}\ \\
\vspace{-2mm}
\includegraphics[width=0.1\linewidth,height=1.2cm]{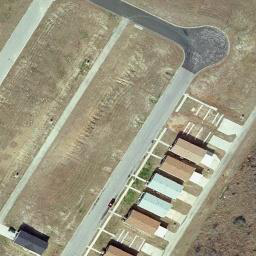}\ &
\includegraphics[width=0.1\linewidth,height=1.2cm]{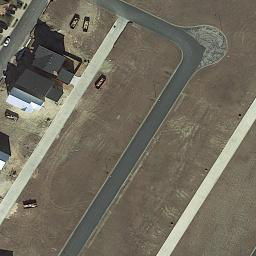}\ &
\includegraphics[width=0.1\linewidth,height=1.2cm]{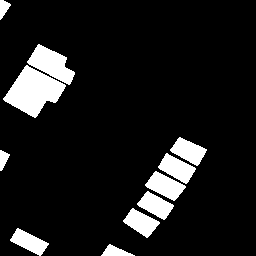}\ &
\includegraphics[width=0.1\linewidth,height=1.2cm]{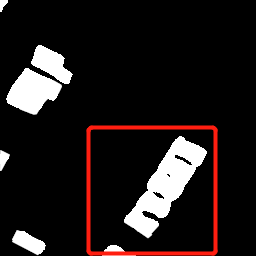}\ &
\includegraphics[width=0.1\linewidth,height=1.2cm]{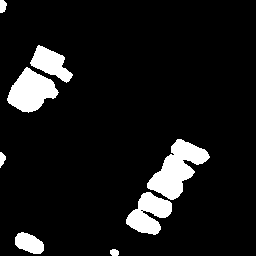}\ &
\includegraphics[width=0.1\linewidth,height=1.2cm]{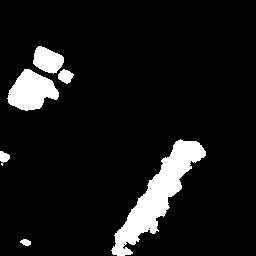}\ &
\includegraphics[width=0.1\linewidth,height=1.2cm]{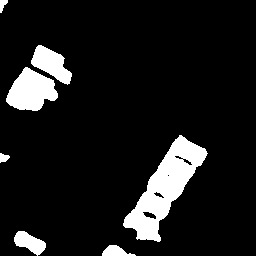}\ \\
 {\tiny T1 Image} & {\tiny T2 Image} & {\tiny Ground Truth} & {\tiny Ours} & {\tiny DTCDSCN} & {\tiny IFNet} & {\tiny CFormer} \ \\
\end{tabular}
}
\caption{Comparison of typical change detection results on more hard and failed samples. Best view by zooming in.}
\label{fig:hard}
\end{figure*}
\begin{figure*}
\resizebox{1\textwidth}{!}
{
\begin{tabular}{@{}c@{}c@{}c@{}c@{}c@{}c@{}c@{}c@{}c@{}c@{}c@{}c}
\includegraphics[width=1\linewidth,height=4.2cm]{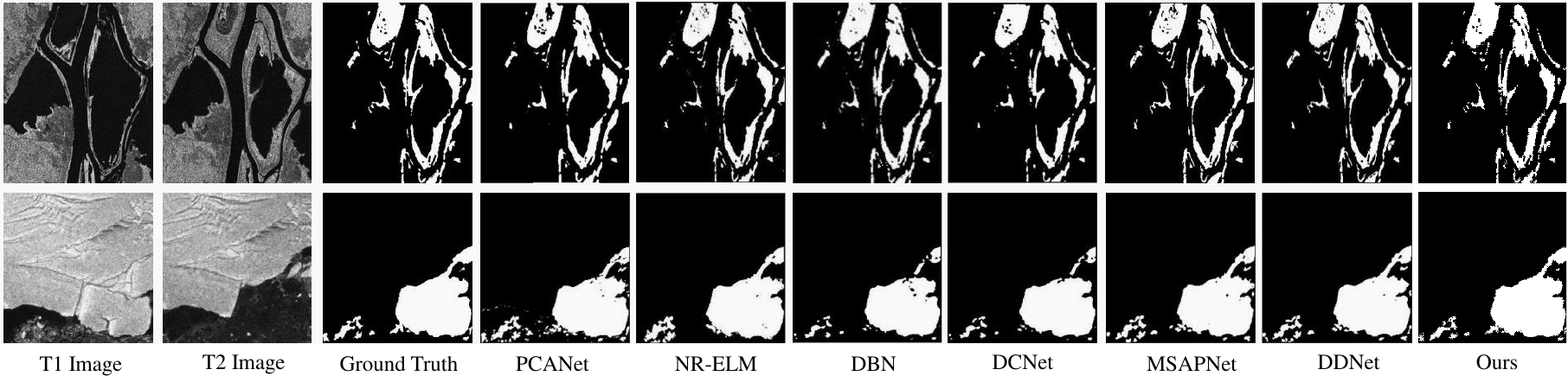}\ \\
\end{tabular}
}
\caption{Comparison of typical change detection results on the Ottawa dataset (top) and Sulzberger dataset (bottom). Images are in disaster environments. Best view by zooming in.}
\label{fig:SAR}
\end{figure*}

To further verify the visual effect, we provide more hard samples and failed results in Fig.~\ref{fig:hard}.
As can be seen, our method performs better than most methods (1st row).
Most of current methods can not detect the two small change regions in the center, while our method can accurately localize them.
Besides, we also show failed examples in the second row of Fig.~\ref{fig:hard}.
As can be seen, all compared methods can not detect all the change regions.
However, our method shows a much more reasonable result than other methods.

To verify the generalization of our method, Fig.~\ref{fig:SAR} shows the change detection results on two SAR image CD datasets.
The other compared methods include the PCANet~\cite{celik2009unsupervised}, NR-ELM~\cite{gao2016change},
DBN~\cite{gong2015change}, DCNet~\cite{gao2019change}, MSAPNet~\cite{wang2020sar} and DDNet~\cite{qu2021change}.
It can be observed that our proposed method shows better results of CD regions and boundaries.
These results also demonstrate the effectiveness of our method on different types of remote sensing images and in disaster environments, including flood and tsunami.
\subsection{More Discussions}
\begin{figure}
\centering
\resizebox{0.48\textwidth}{!}
{
\begin{tabular}{@{}c@{}c@{}}
\includegraphics[width=1\linewidth,height=0.64\linewidth]{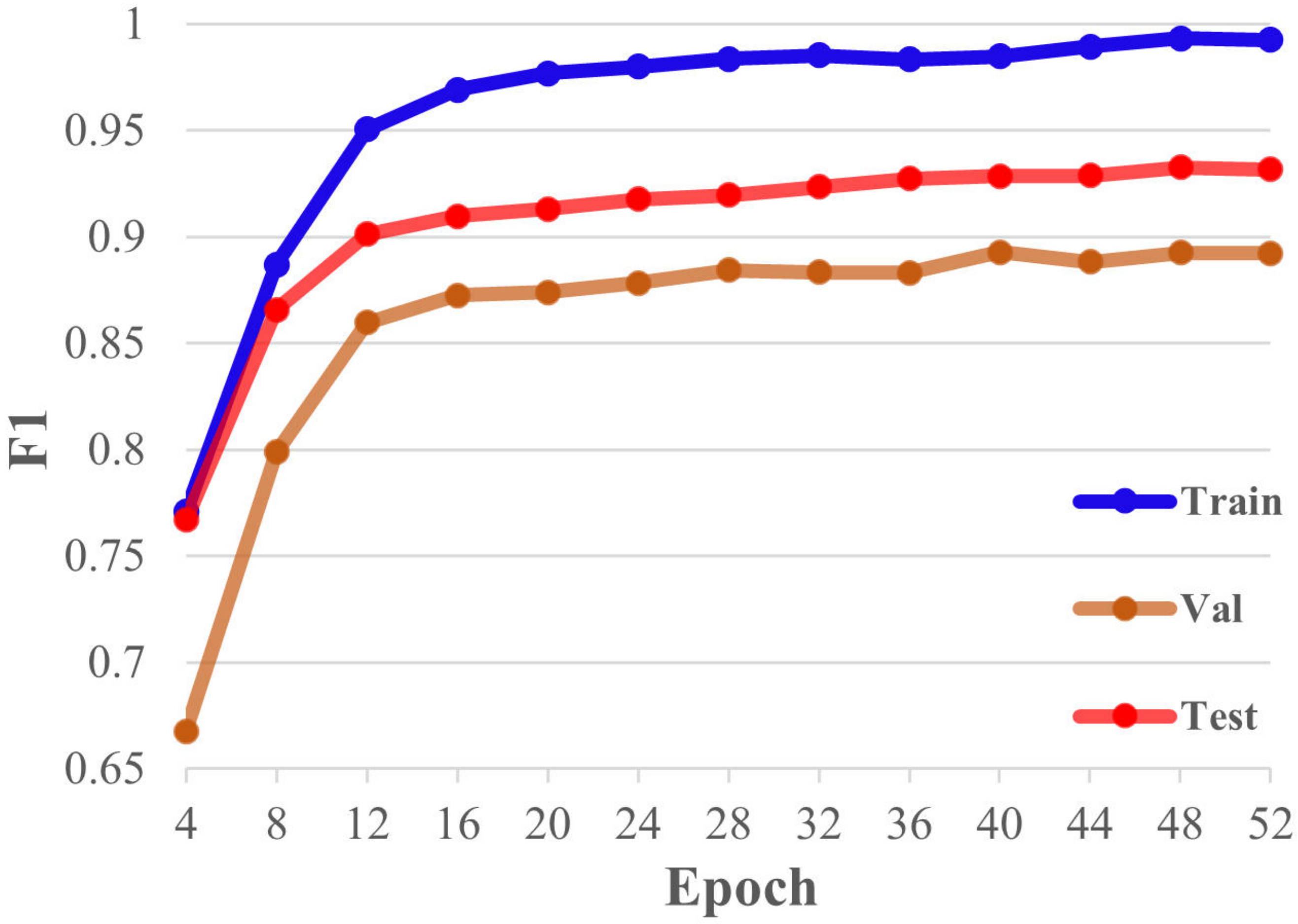} \\
\end{tabular}
}
\caption{Dynamic results of our model on the WHU-CD dataset.}
\label{fig:trainval}
\end{figure}

In order to better clarify the model performance, we plot the dynamic results of our model during the training, validation and testing phases (see Fig.~\ref{fig:trainval}).
It can be observed that our model performs better on the training data than the test data.
This is reasonable and practical since most of current deep learning methods have similar trends.

Previous sections display the performance of changed regions.
To evaluate the performance of regional boundaries, we list the boundary accuracy in Tab.~\ref{table:boundary}.
It can be observed that our method shows much better results than other outstanding methods.
It clearly demonstrates the effectiveness of our proposed method in improving the boundary accuracy.

We note that Transformers (including Fully Transformer Networks) are not new in current remote sensing and computer vision fields.
However, there are some key differences between our work and previous methods:
1) As far as we know, our work is the earliest Transformer-based one, which explicitly handles incomplete regions and irregular boundaries for remote sensing image CD.
2) In our framework, we utilize a Siamese structure to process dual-phase remote sensing images.
Besides, we introduce a pyramid structure to aggregate multi-level visual features from Transformers for feature enhancement.
These designs are totally different from existing works, especially in~\cite{he2022fully} and~\cite{wu2021fully}, which mainly use a simple encoder-decoder+U-Net structure for single image feature extraction.
In~\cite{he2022fully}, the authors utilize a Pyramid Group Transformer (PGT) as the encoder and propose a Feature Pyramid Transformer as the decoder, which is largely based on the typical FPN structure.
Meanwhile, the work in~\cite{wu2021fully} simply stacks Spatial Pyramid Transformers (SPT) imitating the U-Net structure.
Both of them are taking single images as inputs and using an encoder-decoder structure.
While our framework utilizes a Siamese structure to process dual-phase images.
3) We utilize the deeply-supervised learning with multiple boundary-aware loss functions.
These losses are very helpful for more accurate CD.
These facts make our framework more convincing in the CD techniques.
\begin{table}
\setlength{\tabcolsep}{5pt}
\centering
\caption{Boundary accuracies with different methods on LEVIR-CD.}
\label{table:boundary}
\resizebox{0.50\textwidth}{!}
{
\begin{tabular}{|c|c|c|c|c|c|c|c|c|c|c|c|c|c|c|c|c|c|c|c|c|c|c|c|c|c|c|c|c|c|c|c|c|c|c|c|c|c}
\hline
\multicolumn{4}{c|}{Methods}
&\multicolumn{4}{c|}{Ours}&\multicolumn{4}{c|}{ChangeFormer}
&\multicolumn{4}{c|}{BIT}&\multicolumn{4}{c|}{SNUNet}
&\multicolumn{4}{c|}{STANet}&\multicolumn{4}{c}{FC-Diff}
\\
\hline
\multicolumn{4}{c|}{mBA}
&\multicolumn{4}{c|}{71.6}&\multicolumn{4}{c|}{68.7}
&\multicolumn{4}{c|}{65.8}&\multicolumn{4}{c|}{65.2}
&\multicolumn{4}{c|}{63.3}&\multicolumn{4}{c}{60.4}
\\
\hline
\end{tabular}
}
\end{table}
\subsection{Ablation Studies}
In this subsection, we perform extensive ablation studies to verify the effect of key components in our framework.
The experiments are conducted on LEVIR-CD dataset. However, other datasets have similar performance trends.

\textbf{Effects of different Siamese backbones}.
As shown in the 2-3 rows of Tab.~\ref{tab:ablation1}, we introduce the VGGNet-16~\cite{simonyan2014very} and Swin Transformer as the Siamese backbones.
To ensure a fair comparison, we utilize the basic Feature Pyramid (FP) structure~\cite{lin2017feature}.
From the results, one can see that the performance with the Swin Transformer can be consistently improved in terms of Recall, F1, IoU and OA.
The main reason is that the Swin Transformer has a better ability of modeling long-range dependency than VGGNet-16.
\begin{table}[htp]
\setlength{\tabcolsep}{5pt}
\centering
\caption{Performance comparisons with different model variants.}
\resizebox{0.5\textwidth}{!}
{
\begin{tabular}{c|c|c|c|c|cc}
\hline
Models&Pre.&Rec.&F1&IoU&OA\\
\hline
(a) VGGNet-16+FP       &91.98&82.65&87.06&77.09&98.75\\
\hline
(b) SwinT+FP           &91.12&87.42&89.23&80.56&98.91\\
\hline
(c) SwinT+DFE+FP       &91.73&88.43&90.05&81.89&99.00\\
\hline
(d) SwinT+DFE+PCP~\cite{yan2022fully}      &92.71&89.37&91.01&83.51&99.06\\
\hline
(e) SwinT+DFE+PCP      &92.90&89.35&91.90&83.64&99.07\\
\hline
\end{tabular}
}
\label{tab:ablation1}
\end{table}
\begin{figure*}
\centering
\resizebox{1\textwidth}{!}
{
\begin{tabular}{@{}c@{}c@{}c@{}c@{}c@{}c@{}c@{}c@{}c@{}c@{}c@{}c}
\includegraphics[width=0.1\linewidth,height=1.6cm]{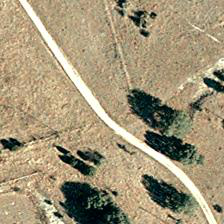}\ &
\includegraphics[width=0.1\linewidth,height=1.6cm]{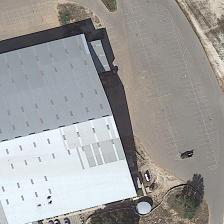}\ &
\includegraphics[width=0.1\linewidth,height=1.6cm]{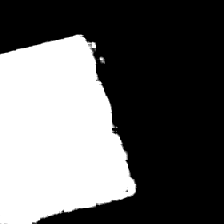}\ &
\includegraphics[width=0.1\linewidth,height=1.6cm]{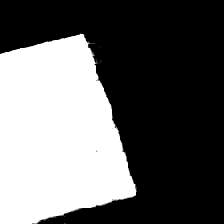}\ &
\includegraphics[width=0.1\linewidth,height=1.6cm]{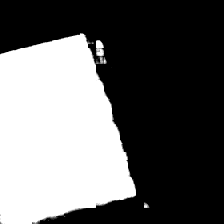}\ &
\includegraphics[width=0.1\linewidth,height=1.6cm]{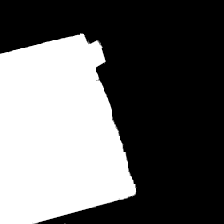}\ &
\includegraphics[width=0.1\linewidth,height=1.6cm]{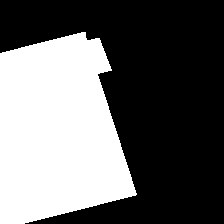}\ \\
 {\tiny T1 Image} & {\tiny T2 Image} & {\tiny Model (b)} & {\tiny Model (c)} & {\tiny Model (d)} & {\tiny Model (e)} & {\tiny Ground Truth}\ \\
\end{tabular}
}
\caption{Visual comparisons of predicted change maps with different models. Best view by zooming in.}
\label{fig:models}
\end{figure*}
\begin{figure*}
\centering
\resizebox{1\textwidth}{!}
{
\begin{tabular}{@{}c@{}c@{}c@{}c@{}c@{}c@{}c@{}c@{}c@{}c@{}c@{}c}
\includegraphics[width=0.1\linewidth,height=1.6cm]{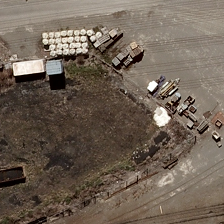}\ &
\includegraphics[width=0.1\linewidth,height=1.6cm]{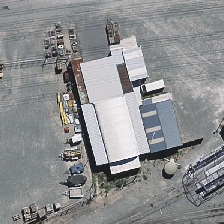}\ &
\includegraphics[width=0.1\linewidth,height=1.6cm]{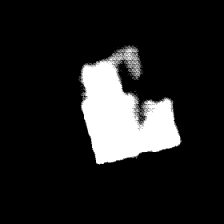}\ &
\includegraphics[width=0.1\linewidth,height=1.6cm]{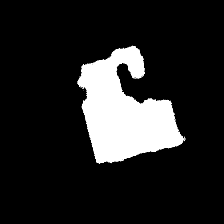}\ &
\includegraphics[width=0.1\linewidth,height=1.6cm]{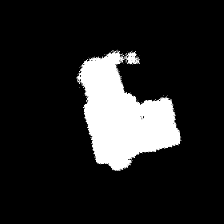}\ &
\includegraphics[width=0.1\linewidth,height=1.6cm]{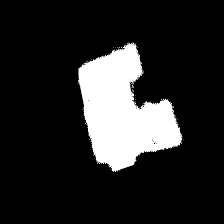}\ &
\includegraphics[width=0.1\linewidth,height=1.6cm]{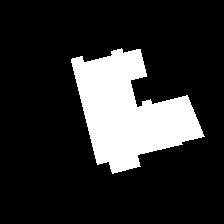}\ \\
 {\tiny T1 Image} & {\tiny T2 Image} & {\tiny BCE} & {\tiny WBCE} & {\tiny WBCE+SSIM} & {\tiny WBCE+SSIM+IOU} & {\tiny Ground Truth}\ \\
\end{tabular}
}
\caption{Visual comparisons of predicted change maps with different losses. Best view by zooming in.}
\label{fig:losses}
\end{figure*}

\textbf{Effects of DFE}.
The fourth row of Tab.~\ref{tab:ablation1} shows the effect of our proposed DFE.
When compared with the $Model (b)~SwinT+FP$, the DFE improves the F1 value from 89.23\% to 90.05\%, and the IoU value from 80.56\% to 81.89\%, respectively.
The main reason is that our DFE considers the temporal information with feature summation and difference, which highlight the change regions.

\textbf{Effects of PCP}.
In order to better detect multi-scale change regions, we introduce the PCP, which is a pyramid structure grafted with a PAM.
If we remove the PAM, the PCP will reduce a basic FP.
We compare it with FP.
From the results in the last two rows of Tab.~\ref{tab:ablation1}, one can see that our PCP achieves a significant improvement in all metrics.
Besides, the improved PCP in this work shows better performances.
Furthermore, adding the PCP also achieves a better visual effect, in which the extracted change regions are complete and the boundaries are regular, as shown in Fig.~\ref{fig:models}.

In addition, we also introduce the Swin Transformer blocks into the PCP as shown in Eq.~\ref{11}.
To verify the effect of different layers, we report the results in Tab.~\ref{tab:ablation2}.
From the results, one can see that the models show better results with equal layers.
The best results can be achieved with $n=4$.
With more layers, the computation is larger and the performance decreases in our framework.
\begin{table}[htp]
\setlength{\tabcolsep}{5pt}
\centering
\caption{Performance comparisons with different decoder layers.}\label{tab:ablation2}
\resizebox{0.42\textwidth}{!}
{
\begin{tabular}{c|c|c|c|c|cc}
\hline
Layers&Pre.&Rec.&F1&IoU&OA\\
\hline
(2,2,2,2)      &91.18&87.00&89.04&80.24&98.90\\
\hline
(4,4,4,4)      &91.65&88.42&90.01&81.83&99.00\\
\hline
(6,6,6,6)      &91.70&88.30&89.96&81.76&98.99\\
\hline
(8,8,8,8)      &91.55&88.47&89.98&81.79&98.99\\
\hline
(2,4,6,8)      &92.13&85.71&88.80&79.86&98.89\\
\hline
\end{tabular}
}
\end{table}

\textbf{Effects of different losses}.
In this work, we introduce multiple losses to improve the CD results.
To show the effects of these losses, we adopt the network structure in~\cite{yan2022fully}.
Tab.~\ref{tab:ablation3} shows the results.
It can be seen that using the WBCE loss can improve the F1 score from 88.75\% to 90.01\% and the IoU from 79.78\% to 81.83\%.
Using the SSIM loss achieves the F1 score of 90.11\% and the IoU of 82.27\%.
Using the SIoU loss achieves the F1 score of 91.01\% and the IoU of 83.51\%.
\begin{table}[htp]
\setlength{\tabcolsep}{5pt}
\centering
\caption{Performance comparisons with different losses.}
\label{tab:ablation3}
\resizebox{0.48\textwidth}{!}
{
\begin{tabular}{c|c|c|c|c|cc}
\hline
Losses&Pre.&Rec.&F1&IoU&OA\\
\hline
BCE            &90.68&86.91&88.75&79.78&98.88\\
\hline
WBCE           &91.65&88.42&90.01&81.83&99.00\\
\hline
WBCE+SSIM      &91.71&88.57&90.11&82.27&99.01\\
\hline
WBCE+SSIM+SIoU &92.71&89.37&91.01&83.51&99.06\\
\hline
\end{tabular}
}
\end{table}
\begin{table}
\setlength{\tabcolsep}{5pt}
\centering
\caption{Performances with different input resolutions on LEVIR-CD.}
\label{table:resolutions}
\resizebox{0.50\textwidth}{!}
{
\begin{tabular}{|c|c|c|c|c|c|c|c|c|c|c|c|c|c|c|c|c|c|c|c|c|c|c|c|c|c|c|c|c|c|c|c|c|c|c|c|c|c|c|c|}
\hline
\multicolumn{4}{c|}{Resolution}
&\multicolumn{4}{c}{Pre.}&\multicolumn{4}{c}{Rec.}
&\multicolumn{4}{c}{F1}&\multicolumn{4}{c}{IoU}
&\multicolumn{4}{c|}{OA}&\multicolumn{4}{c}{Flops(G)}
\\
\hline
\multicolumn{4}{c|}{256$\times$256}
&\multicolumn{4}{c}{92.90}&\multicolumn{4}{c}{89.35}
&\multicolumn{4}{c}{91.90}&\multicolumn{4}{c}{83.64}
&\multicolumn{4}{c|}{99.07}&\multicolumn{4}{c}{48}
\\
\multicolumn{4}{c|}{384$\times$384}
&\multicolumn{4}{c}{93.01}&\multicolumn{4}{c}{90.11}
&\multicolumn{4}{c}{92.12}&\multicolumn{4}{c}{84.20}
&\multicolumn{4}{c|}{99.10}&\multicolumn{4}{c}{151}
\\
\multicolumn{4}{c|}{512$\times$512}
&\multicolumn{4}{c}{93.23}&\multicolumn{4}{c}{90.35}
&\multicolumn{4}{c}{92.25}&\multicolumn{4}{c}{85.10}
&\multicolumn{4}{c|}{99.32}&\multicolumn{4}{c}{201}
\\
\hline
\end{tabular}
}
\end{table}

We also display some typical examples for the visual effects, as shown in Fig.~\ref{fig:losses}.
From the results, one can see that using the WBCE loss can help the model focus on the most change regions.
With the SSIM loss, the framework can improve the structural information of the change regions.
Using the SIoU loss can ensure the global completeness.
As a result, combining all of them can achieve the best results, which proves the effectiveness of all loss terms.
This fact is consistent with the quantitative results in Tab.~\ref{tab:ablation3}.

\textbf{Scaling to higher resolutions.} Remote sensing images always hold large resolutions.
The resolution concern is very valuable.
In fact, our work can process a higher resolution with SwinT-Base/Small/Tiny.
However, we adopt a low resolution (256$\times$256), mainly considering the fairness.
Most of compared methods utilize cropping for generating input images.
Thus, we follow them and realize fair comparisons.
Tab.~\ref{table:resolutions} shows the performance analysis with different resolutions on LEVIR-CD.
One can see that our method can naturally scale to higher resolutions and show slightly better results.
\section{Conclusion and Future Work}
In this work, we propose a new learning framework named TranY-Net for change detection of dual-phase remote sensing images.
It improves the feature extraction from a global view and combines multi-level visual features in a pyramid manner.
Technically, we first utilizes a Siamese network with the pre-trained Swin Transformers to extract long-range dependency information.
Then, we introduce a pyramid structure to aggregate multi-level visual features, improving the feature representation ability.
Finally, we utilize the deeply-supervised learning with multiple loss functions for model training.
Extensive experiments on four public CD benchmarks demonstrate that our proposed framework shows better performances than most state-of-the-art methods.
However, our methods have some shortcomings, such as high computation, the need of image dense labeling, etc.
In future works, we will explore more efficient structures of Transformers to reduce the computation.
We will also develop unsupervised or weakly-supervised methods to relieve the burden of remote sensing image labeling.
In addition, since our method takes dual-phase images, it can be easily used for other similar multi-modal/temporal tasks, such as RGB-D/T image segmentation, MRI-CT image fusion, video segmentation, etc.
We will verify them in the computer vision field.

\ifCLASSOPTIONcaptionsoff
  \newpage
\fi
\bibliographystyle{IEEEtran}
\bibliography{IEEEabrv,refs}
\end{document}